\documentclass[10pt,journal,compsoc,twocolumn]{IEEEtran}

\usepackage{booktabs}
\usepackage{multirow}
\usepackage{flushend}
\usepackage{graphicx}
\usepackage{rotating}
\usepackage{hyperref}
\usepackage{dsfont,amsfonts,amssymb,amsmath,color}
\usepackage{latexsym}
\usepackage{amsmath}
\usepackage{amsfonts}
\usepackage{algorithm}
\usepackage[noend]{algpseudocode}
\usepackage{graphicx}
\usepackage{color}
\usepackage{array}
\usepackage{algorithm}  
\usepackage{algpseudocode}
\usepackage{amsmath}  
\usepackage{makecell}
\usepackage{cases}
\usepackage{color}
\usepackage{amsthm}

\usepackage{soul}
\usepackage{tabularx}
\usepackage{makecell}
\usepackage{mwe}
\usepackage{subfig}
\usepackage{fixltx2e}

\newtheorem{define}{Definition}[section]
\newcommand\FIXED[1]{\textcolor{black}{#1}}

\makeatletter
\def\hlinewd#1{%
  \noalign{\ifnum0=`}\fi\hrule \@height #1 \futurelet
   \reserved@a\@xhline}
\makeatother

\begin{document}

\title{Higher-Order Attribute-Enhancing Heterogeneous Graph Neural Networks}

\author{Jianxin Li,
        ~Hao Peng,
        ~Yuwei Cao,
        ~Yingtong Dou,
        ~Hekai Zhang,
        ~Philip S. ~Yu~\IEEEmembership{Fellow,~IEEE,}
        ~Lifang He\\
\IEEEcompsocitemizethanks{
\IEEEcompsocthanksitem Jianxin Li and Hao Peng are with Beijing Advanced Innovation Center for Big Data and Brain Computing, Beihang University, Beijing 100083, China. E-mail: \{lijx, penghao\}@act.buaa.edu.cn.
\IEEEcompsocthanksitem Yuwei Cao, Yingtong Dou, and Philip S. Yu are with the Department of Computer Science, University of Illinois at Chicago, Chicago, IL 60607, USA. E-mail:\{ycao43, ydou5, psyu\}@.uic.edu.
\IEEEcompsocthanksitem Hekai Zhang is with School of Information Science and Engineering, Yanshan University, Qinhuangdao 066004, China. E-mail: hekai\_zhang@163.com.
\IEEEcompsocthanksitem Lifang He is with the Department of Computer Science and Engineering, Lehigh University, Bethlehem, PA 18015 USA. E-mail: lih319@lehigh.edu.
}
\thanks{Manuscript received April, 2021. (Corresponding author: Hao Peng and Jianxin Li.)}
}

\markboth{}%
{Shell \MakeLowercase{\textit{et al.}}: Bare Demo of IEEEtran.cls for Computer Society Journals}

\IEEEtitleabstractindextext{%
\begin{abstract}
Graph neural networks (GNNs) have been widely used in deep learning on graphs.
They can learn effective node representations that achieve superior performances in graph analysis tasks such as node classification and node clustering.
However, most methods ignore the heterogeneity in real-world graphs.
Methods designed for heterogeneous graphs, on the other hand, fail to learn complex semantic representations because they only use meta-paths instead of meta-graphs.
Furthermore, they cannot fully capture the content-based correlations between nodes, as they either do not use the self-attention mechanism or only use it to consider the immediate neighbors of each node, ignoring the higher-order neighbors.
We propose a novel Higher-order Attribute-Enhancing (HAE) framework that enhances node embedding in a layer-by-layer manner. 
Under the HAE framework, we propose a Higher-order Attribute-Enhancing Graph Neural Network (HAE\textsubscript{GNN}) for heterogeneous network representation learning.
HAE\textsubscript{GNN} simultaneously incorporates meta-paths and meta-graphs for rich, heterogeneous semantics, and leverages the self-attention mechanism to explore content-based nodes' interactions.
The unique higher-order architecture of HAE\textsubscript{GNN} allows examining the first-order as well as higher-order neighborhoods. 
Moreover, HAE\textsubscript{GNN} shows good explainability as it learns the importances of different meta-paths and meta-graphs.
HAE\textsubscript{GNN} is also memory-efficient, for it avoids per meta-path based matrix calculation. 
Experimental results not only show HAE\textsubscript{GNN}'s superior performance against the state-of-the-art methods in node classification, node clustering, and visualization, but also demonstrate its superiorities in terms of memory efficiency and explainability.
\end{abstract}

}


\maketitle

\IEEEdisplaynontitleabstractindextext

\IEEEpeerreviewmaketitle

\IEEEraisesectionheading{\section{Introduction}\label{sec:introduction}}
Networks are ubiquitous in our daily lives with examples such as transportation networks and social networks.
Network representation learning, as the foundation of downstream analysis tasks including node classification~\cite{bhagat2011node,hamilton2017inductive,velivckovic2017graph}, node clustering~\cite{velivckovic2017graph} and visualization~\cite{maaten2008visualizing}, has been attracting increasing attention in both academia and industry. 
In recent years, Graph Neural Networks (GNNs) have been widely adopted in network representation learning, and have shown state-of-the-art performances. 
For example, \cite{kipf2016semi} proposes a Graph Convolutional Network (GCN) framework, which is a variant of convolutional neural networks. 
It operates directly on graphs and incorporates the 1-step neighborhood for each node's embedding. 
Graph Attention Network (GAT)~\cite{velivckovic2017graph} further introduces a self-attention mechanism~\cite{bahdanau2014neural} to focus on the most relevant neighboring nodes. 
These methods~\cite{kipf2016semi,wang2016structural,velivckovic2017graph} treat networks as homogeneous ones, while there are some most recent studies~\cite{wang2019heterogeneous,yun2019graph,zhang2019heterogeneous,hu2020heterogeneous,fu2020magnn} that also consider the heterogeneity of real-world networks. 
Heterogeneous Graph Attention Network (HAN)~\cite{wang2019heterogeneous} learns trainable weights to fuse meta-path sampling based node embeddings.
Heterogeneous Graph Neural Network (HetGNN)~\cite{zhang2019heterogeneous} aggregates multi-modal features from heterogeneous neighbors by combining bi-LSTM, self-attention, types combination and other complex neural network technologies.
Heterogeneous Graph Transformer (HGT)~\cite{hu2020heterogeneous} leverages type-dependent parameters based on mutual attention, message passing, residual connection, target-specific aggregation function, etc.
Metapath Aggregated Graph Neural Network (MAGNN)~\cite{fu2020magnn} makes use of meta-path sampling, intra-metapath and inter-metapath aggregation technologies to generate the embedding of the target type node.
Graph Transformer Network (GTN)~\cite{yun2019graph} learns task-oriented meta-paths discovering.
However, the heterogeneity issue has not been fully addressed in a convenient manner, as the existing methods either limit their sources of semantics to meta-paths that cannot fully capture the semantic-based similarities between nodes, or employ complex neural network technologies.

\begin{figure*}
    \centering
    \includegraphics[width = 16cm]{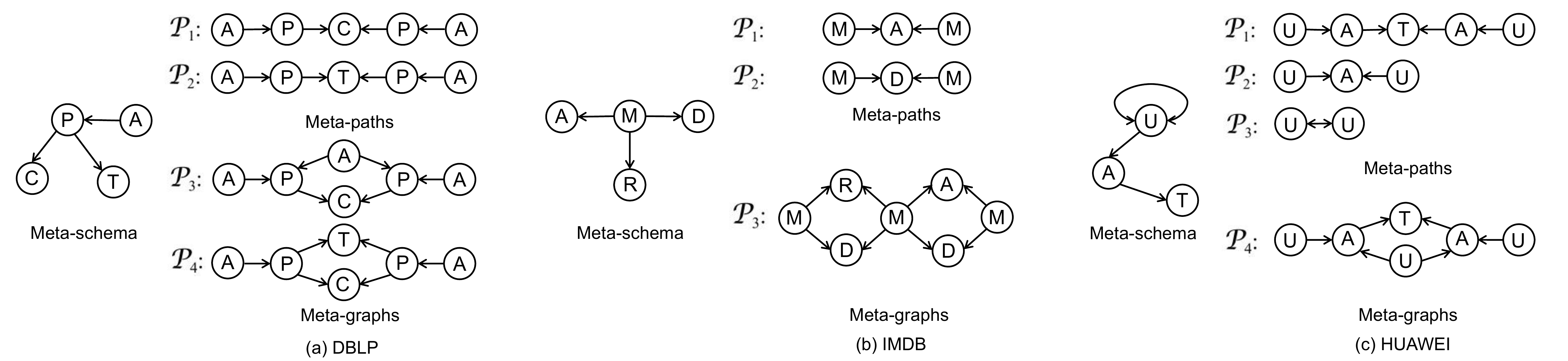}
    \caption{Meta-schema, meta-paths and meta-graphs of the DBLP, IMDB and HUAWEI datasets. In (a), A, P, C, and T stand for \textit{Author}, \textit{Paper}, \textit{Conference}, and \textit{Term}, respectively. In (b), A, M, D, and R stand for \textit{Actor}, \textit{Movie}, \textit{Director}, and \textit{Rating}, respectively. In (c), U, A, and T stand for \textit{User}, \textit{Application}, and \textit{Application type}, respectively.}
    \label{meta}
\end{figure*}

In this work, we consider a Heterogeneous Information Network (HIN)~\cite{shi2016survey} - i.e., a network in which there are different types of nodes and edges involved. 
For capturing the complex semantics in a HIN, the concepts of meta-schema, as well as the meta-paths and meta-graphs defined at the schema level, have been developed. 
Both meta-paths and meta-graphs can reflect the similarity between two nodes; however, it has been shown by studies on recommendation system~\cite{zhao2017meta}, attributed graph representation learning~\cite{sankar2019meta} and unsupervised network embedding~\cite{he2019hetespaceywalk,zhang2018metagraph2vec} that compared to meta-paths, meta-graphs can capture more complex semantics.
Take the citation network DBLP shown in Fig.~\ref{meta}(a) as an example, meta-path $P_1$ decides that two authors are similar as long as they publish their works in the same conference, while meta-graph $P_3$ further requires them to co-work with the same third author in order for these two authors to be considered as similar.
Therefore, instead of exploring the semantics solely by meta-paths, incorporating meta-graphs is essentially a better way to characterize the rich semantics~\cite{yang2018meta}.
In fact, experimental results in~\cite{sun2018joint} also suggest that combined use of meta-paths and meta-graphs better help capture the similarities between entities.
It is worth noting that when using multiple meta-paths and meta-graphs in combination, we should put different emphases on each of them, as some meta-paths or meta-graphs may convey more important semantics compared to the others. 
Also, since different meta-paths and meta-graphs learn embedding in different semantic aspects, figuring out an effective way to assemble them is of crucial importance.

Another issue with the current methods is that they all ignore the content-based relatedness between nodes to some extent.
To the best of our knowledge, none of the current methods is capable of capturing interdependencies between nodes and their higher-order neighbors.
Given a node, we can define its neighbors using meta-paths and meta-graphs. 
According to the self-attention mechanism \cite{bahdanau2014neural}, a node and a neighbor are related, and the extent of their relatedness, i.e., their content-based similarity \cite{cordonnier2019relationship}, can be measured by an attention score that computes the distance of their representations. 
However, two nodes that are non-immediate neighbors of each other (i.e., not directly connected by any meta-paths or meta-graphs, but share some common meta-paths or meta-graphs based neighbors) are also correlated, and we refer to such nodes as higher-order neighbors. 
For example, in the DBLP dataset, two authors $a_1$ and $a_2$ may publish their works in one conference $c_1$, while $a_1$ and a third author $a_3$ might have their works published in the other conference $c_2$.
Then, according to meta-path $P_1$, $a_2$ and $a_3$ are the neighbors of $a_1$. 
In such a case, although $a_2$ and $a_3$ are not the direct neighbors of each other, they are still correlated, since both of them are related to $a_1$. 
The higher-order neighbors should not be ignored when considering the content-based interdependencies between nodes. 
Therefore, a self-attention structure that simultaneously examines the first-order, second-order, as well as higher-order neighborhoods, is desired.

In this paper, we propose a novel Higher-order Attribute-Enhancing Graph Neural Network (HAE$_{GNN}$) to address the above issues. 
Specifically, we design a Higher-order Attribute-Enhancing framework (HAE) to combine multiple Semantic-based Convolutional Layers (SCLs) and Content-based self-Attention Layers (CALs). 
We refer to SCLs and CALs collectively as \textit{attribute-enhancing layers}. 
We elaborate on these three components as follows.

\textbf{Semantic-Based Convolutional Layer (SCL).} 
SCL enhances the input node features with semantics. 
It incorporates multiple meta-paths and meta-graphs to capture the rich, heterogeneous semantics in them. 
These meta-paths and meta-graphs need to begin and end with the same node type (referred to as the \textit{target type}). 
Each meta-path/meta-graph is associated with a trainable weight. 
The weights allow us to focus more on the most important meta-paths/meta-graphs.
Based on that, we define a meta-paths/meta-graphs instances based similarity between two nodes of the target type, and this similarity is parameterized by the weights of meta-paths/meta-graphs. 
After that, we construct an adjacency matrix composed of the similarities between all pairs of target nodes. 
At this point, the raw heterogeneous network has been converted into a homogeneous network with weighted connections between the target nodes. 
We then apply graph convolutions~\cite{kipf2016semi} on the converted network. 
In this way, SCL fuses the input node features with meaningful semantics.

\textbf{Content-Based Self-Attention Layer (CAL).} 
CAL leverages the content-based interactions between nodes and their neighbors for enhanced node embedding. 
We construct a binary adjacency matrix based on the meta-paths/meta-graphs based connections between nodes. 
Similar to the SCL case, the adjacency matrix defines a converted, homogeneous network.
Given the converted network, we then leverage the self-attention mechanism~\cite{bahdanau2014neural} to compute the hidden representations of each node by paying attention to its neighbors. 
Here, the term content-based attention is used as opposed to position-based attention \cite{cordonnier2019relationship}. 
Namely, the attention scores depend on the representations of node-neighbor pairs, and we do not differentiate the order of the neighbors when applying attention.

\textbf{Higher-Order Attribute-Enhancing Framework (HAE).} 
The HAE framework repeatedly stacks multiple SCLs and CALs. 
In this way, HAE gradually enhances the initial input node features with semantics and content-based interactions, layer by layer.
Especially, by stacking multiple CALs, we can deploy higher-order neighbors. 
It is worth noting that the HAE framework allows the stacking of an arbitrary number of SCLs and CALs in an arbitrary sequence. 
This provides more flexibility, as the user can organize the building blocks according to the needs of the task.
HAE$_{GNN}$, specifically, is composed of one SCL followed by a number of CALs. 
We adopt such a design since it has been shown by studies in computer vision (CV)~\cite{parmar2019stand} that self-attention is especially impactful when used in later layers, and after the convolutional layers.
In this study, we also experiment on more variations under the HAE framework for comparison.

We conduct extensive experiments on three datasets to evaluate the proposed HAE$_{GNN}$ model and compare it with the state-of-the-art baselines. 
The results show that HAE$_{GNN}$ outperforms the baselines. 
Besides, HAE$_{GNN}$ demonstrates superiorities in terms of efficiency and explainability. 
We also present experimental results to illustrate the effectiveness of the proposed semantic capturing approach as well as higher-order architecture.
The code of this work is publicly available at \url{https://github.com/RingBDStack/HAE}.

The main contributions of this paper can be summarized as follows:
1) We propose a novel HAE$_{GNN}$ framework for heterogeneous network representation learning. 
It incorporates meta-paths and meta-graphs for semantics, while leveraging the self-attention mechanism to explore content-based interactions between nodes and their immediate and higher-order neighbors.
2) Compared to the existing methods, HAE$_{GNN}$ not only incorporates richer heterogeneous semantics, but also better captures the long-range content-based interdependencies of nodes.
3) Extensive evaluations on three datasets demonstrate the effectiveness of the proposed HAE$_{GNN}$ over strong baselines in tasks including node classification, node clustering, and visualization.
4) Experimental results show that HAE$_{GNN}$ is memory-efficient. 
Compared to the best baselines, HAE$_{GNN}$ is able to give better results with lower memory consumption.
5) Experimental results also demonstrate the explainability of our semantic meta-paths and meth-graphs.


\section{Related work}\label{sec:related_work}
Our work is closely related to four topics, i.e., graph neural networks, network representation learning, and the available approaches to incorporate semantics as well as self-attention. 
We describe details as follows.

\textbf{Graph Neural Networks.} 
GNNs, introduced by~\cite{gori2005new,scarselli2008graph}, are extended from recursive neural networks so as to operate on graph-structured data. 
In recent years, there is an increasing number of GNNs that generalize convolutional operation~\cite{velivckovic2017graph}.
Such GNNs can be divided into two classes, i.e., spectral ones~\cite{bruna2013spectral,henaff2015deep,defferrard2016convolutional,kipf2016semi,velivckovic2017graph} which compute a spectral representation of the graphs, and non-spectral ones~\cite{duvenaud2015convolutional,atwood2016diffusion,hamilton2017inductive} which define convolutions directly on the graph to operate on groups of spatially close neighbors. 
Compared to non-spectral approaches, spectral approaches naturally deal with different sized neighborhoods~\cite{velivckovic2017graph}. 
For example, \cite{kipf2016semi} defines a simplified convolution that operates in a 1-step neighborhood around each node. 
\cite{velivckovic2017graph}, which is also a spectral approach, improves performance by leveraging masked self-attention to capture the content-based interactions between nodes and their 1-step neighbors. 
Despite their successes in network analysis tasks, \cite{velivckovic2017graph} and \cite{kipf2016semi} neglect the heterogeneous semantics in data.

\textbf{Network Representation Learning.} 
This class of methods learns low dimensional vector representations for nodes in the networks.
The derived representations are usually called node embedding, which can be applied to downstream network analysis tasks~\cite{cui2018survey}. 
Network representation learning methods can be divided into unsupervised and semi-supervised ones.
Unsupervised techniques preserve the structural information of the network without any prior labeling knowledge. 
Among unsupervised techniques, there are several methods proposed for homogeneous networks, such as random walk based ones~\cite{grover2016node2vec,perozzi2014deepwalk}, matrix factorization based ones~\cite{ou2016asymmetric,wang2017community}, topological structure and attribute proximity based ones~\cite{gao2018deep}, and others~\cite{tang2015line}.
In addition, recent research work of unsupervised high-order network representation learning models~\cite{RossiHigherOrder2018,carranza2020higher} have shown the importance of motif, typed-motif or graphlet in long-distance feature representation.
NEU algorithm~\cite{yang2017fast} considers first-order proximity and second-order proximity, and formalizes proximity matrix construction and dimension reduction to learn unsupervised high-order network embedding.
There are also heterogeneous methods that incorporate semantics, i.e., heterogeneous structural information.
For example, Metapath2vec~\cite{dong2017metapath2vec} explores semantics contained in meta-paths through meta-path guided random walking. 
HeteSpaceyWalk~\cite{he2019hetespaceywalk} formalizes the meta-path guided random walk as a higher-order Markov Chain Process to attain stationary distribution among nodes and extends its method so as to be applied on meta-graphs and meta-schemas.
Our works pay more attention to how to make use of multiple layers of graph neural networks to learn higher-order heterogeneous graph representation.

Instead of preserving solely the structural information, semi-supervised network representation learning methods based on GNNs further leverage available labels to tailor the embedding for specific tasks~\cite{yun2019graph}.
Many semi-supervised methods such as the aforementioned~\cite{kipf2016semi,wang2016structural,velivckovic2017graph,xu2019powerful}, although introducing techniques such as spectral convolution~\cite{bruna2013spectral} and attention mechanism~\cite{bahdanau2014neural} for better performances, share the common limitation of ignoring the heterogeneity in real-world networks. 
SNDE~\cite{wang2016structural} is a homogeneous model that learns both supervised local network structure and unsupervised global network structure.
GTN~\cite{yun2019graph} addresses the heterogeneity issue by learning weighted, task-oriented meta-paths. 
However, it neglects the content-based nodes' interactions for not utilizing the attention mechanism~\cite{bahdanau2014neural} in its design. 
HAN~\cite{wang2019heterogeneous} includes both semantic-level attention and node-level attention. 
Nevertheless, \cite{wang2019heterogeneous} does not incorporate meta-graphs, and its semantics come only from meta-paths. 
Furthermore, \cite{wang2019heterogeneous} does not consider the long-range dependencies between nodes, and its node-level attention explores solely the first-order neighborhoods. 
HetGNN~\cite{zhang2019heterogeneous} conducts neighbors sampling to capture the interactions between strongly correlated heterogeneous neighbors; however, the semantically meaningful meta-paths and meta-graphs are not leveraged.
HGT~\cite{hu2020heterogeneous} implements complex neural aggregations among heterogeneous neighbors, and gives up semantically meaningful meta-paths and meta-graphs.
RSHN~\cite{zhu2019relation} utilizes graph structure and implicit relation structural information to simultaneously learn node and edge type embedding.
MAGNN~\cite{fu2020magnn} explores meta-path sampling based intra-metapath and inter-metapath aggregations among heterogeneous neighbors, and ignores more semantical meta-graphs.
Overall, existing semi-supervised methods incorporate insufficient semantics, and cannot fully leverage the content-based relatedness between nodes in a convenient manner.

\textbf{Incorporate Semantics.} 
Studies that deal with heterogeneous graphs need to properly address the heterogeneity of data. 
Examples can be found in unsupervised~\cite{dong2017metapath2vec,zhang2018metagraph2vec,he2019hetespaceywalk,peng2021lime} and semi-supervised~\cite{wang2019heterogeneous,zhang2019heterogeneous,zhu2019relation,hu2020heterogeneous,fu2020magnn} network embedding, as well as in other fields such as event mining~\cite{peng2019fine,peng2021streaming}, medical data analysis~\cite{cao2020multi}, cyber threat intelligence~\cite{gao2020hincti}, etc.
All of the above studies adopt the most common approach: they first model the data as HINs~\cite{shi2016survey}, then leverage meta-paths and/or meta-graphs. 
Studies on recommendation~\cite{zhao2017meta}, semi-supervised attributed graph embedding~\cite{sankar2019meta} and unsupervised network embedding~\cite{he2019hetespaceywalk} have shown that compared to meta-paths, meta-graphs can capture more complex semantics, and combined use of both gives better experimental results. 
Our model simultaneously leverages meta-paths and meta-graphs, and therefore incorporates richer semantics compared to previous state-of-the-art semi-supervised network embedding methods~\cite{wang2019heterogeneous,fu2020magnn}. 
Recently developed content-aware heterogeneous network representation technologies demonstrate superior performance in downstream applications.
For example, task-guided combination model~\cite{chen2017task} utilizes useful meta-paths selected by specific tasks, e.g., the author identification, and human experience to help generate target types of node embedding in an unsupervised manner.
Camel~\cite{zhang2018camel} combines textual content semantic and meta-path augmented structure features to learn node embedding in a supervised manner.
SHINE~\cite{wang2018shine} learns users’ embedded representation from multiple heterogeneous networks, including sentiment network, social network, and profile network, in a supervised manner.
Our model presents a more generalized and deep content-aware heterogeneous graph representation in a semi-supervised manner.
Embedding learned from different meta-paths/meta-graphs are in different semantic aspects and need to be assembled properly. 
\cite{yun2019graph} concatenates the embedding together.
\cite{wang2019heterogeneous} maps all embedding into one same space then computes a weighted aggregation. 
In contrast, the SCL in our model fuses all meta-paths and meta-graphs before learning. 
In this way, our model naturally learns embedding in one vector space and saves the trouble of further transformations and concatenations.

\textbf{Incorporate Self-Attention.} 
The attention mechanism, since its introduction~\cite{bahdanau2014neural}, has enjoyed rich success in various Natural Language Processing (NLP)~\cite{vaswani2017attention,peng2019hierarchical} and CV~\cite{hu2018squeeze,tan2019mnasnet} tasks.
Compared to convolution, the attention mechanism better handles long-range dependencies~\cite{xu2019powerful}. 
In particular, the self-attention mechanism captures the content-based similarity between two entities by computing an attention score that measures the distance of their representations~\cite{cordonnier2019relationship}.

Existing heterogeneous network representation learning methods~\cite{wang2019heterogeneous,yun2019graph,zhang2019heterogeneous,hu2020heterogeneous,fu2020magnn,zhu2019relation} are not using the self-attention mechanism to its maximum extent in a convenient way.
Specifically, \cite{yun2019graph}, \cite{zhang2019heterogeneous} and \cite{zhu2019relation} don't leverage node-level self-attention, and ignore the content-based interactions between nodes (though \cite{zhang2019heterogeneous} leverages type-level attention to combine the type-wise neighbors' embedding by using bi-LSTM units). 
\cite{hu2020heterogeneous} employs complex mutual attention and message passing technologies, and \cite{fu2020magnn} develops both intra-metapath and inter-metapath aggregation technologies.
It is worthwhile to develop a convenient new framework for heterogeneous graph representation learning that considers both node-level self-attention and content-based interactions.
The node-level attention architecture in \cite{wang2019heterogeneous} is single-layered and considers solely the first-order neighbors of each node. 
In contrast, our model adopts a higher-order architecture that stacks multiple CALs. 
In this way, our model pays attention to the first-order as well as higher-order neighborhoods and fully captures content-based interdependencies between nodes. 
The effectiveness of such higher-order self-attention architecture has been proven by the latest studies in CV~\cite{parmar2019stand}. 
Moreover, our model combines convolutional (through SCLs) with self-attention (through CALs) for better performance. 
This approach is also shown to be useful by most recent CV studies~\cite{tan2019mnasnet} that reach state-of-the-art performances through combining attention with convolutional features.

\begin{figure*}
    \centering
    \includegraphics[width = 18cm]{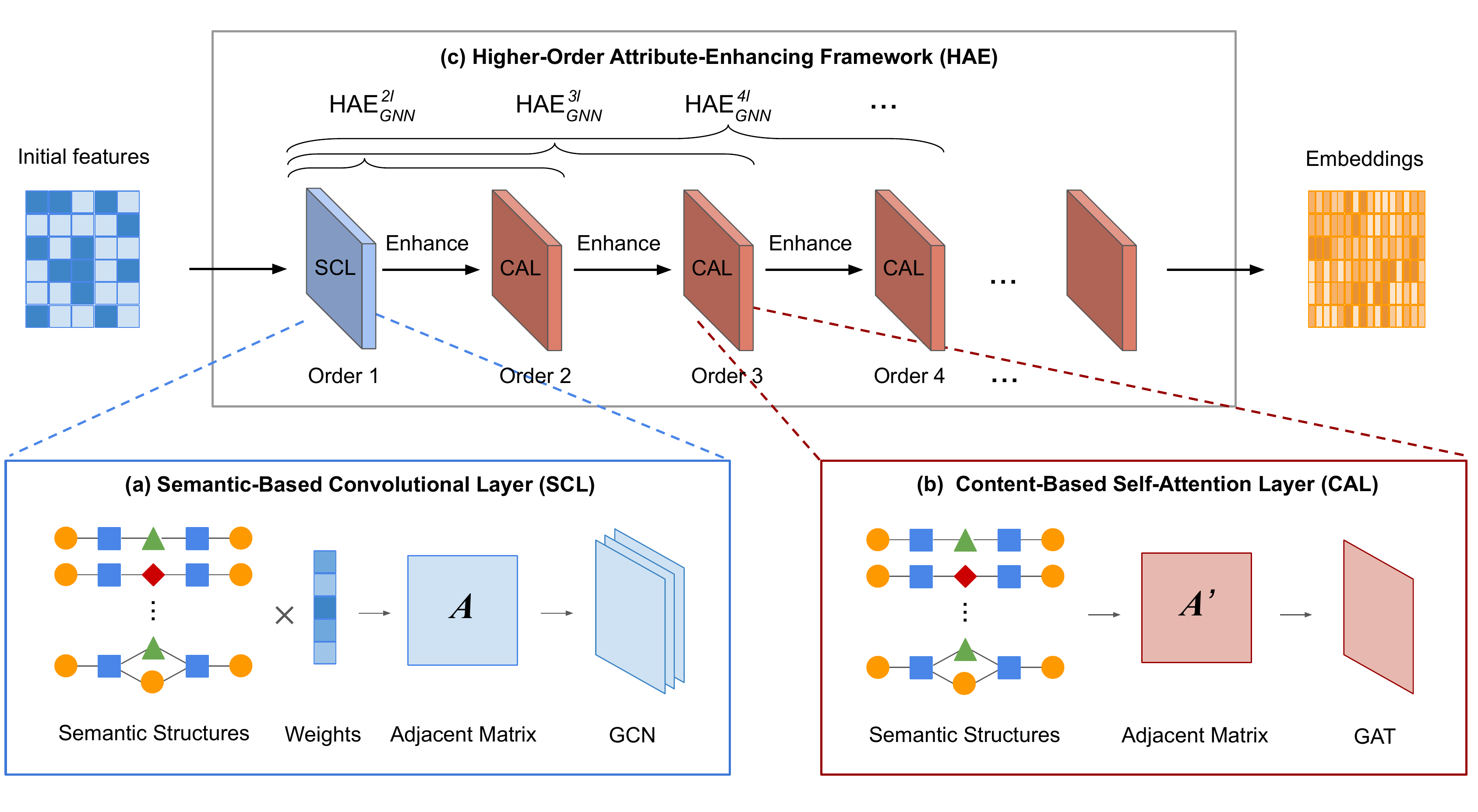}
    \caption{The architecture of the proposed Higher-order Attribute-Enhancing Graph Neural Network (HAE$_{GNN}$). HAE$_{GNN}$ contains three components, i.e., (a) Semantic-Based Convolutional Layer (SCL), (b) Content-Based Self-Attention Layer (CAL), and (c) Higher-Order Attribute-Enhancing Framework (HAE). The inputs of the model are initial features such as the BOW representations of the keywords related to the nodes.}
    \label{framework}
\end{figure*}

\section{Preliminary}\label{sec:preliminary}
We present the definitions of concepts related to this work, including heterogeneous information network, meta-schema, meta-path and meta-graph.

\begin{define}{\textbf{Heterogeneous Information Network}~\cite{shi2016survey}}
A heterogeneous information network (HIN) is a graph $G = (\mathcal{V},\mathcal{E})$, where $\mathcal{V}$ and $\mathcal{E}$ stand for collections of nodes and edges that are of various types.
\end{define}
For example, the DBLP dataset used in this study is a HIN. 
It contains nodes of multiple types, including authors, papers, conferences and terms (keywords). 
It also contains edges (relations) of different types such as \textit{publish} relation between authors and papers, \textit{published in} relation between papers and conferences, and \textit{contains} relation between papers and terms.

\begin{define}{\textbf{Meta-Schema}~\cite{shi2016survey}.}
Given a HIN $G = (\mathcal{V},\mathcal{E})$, its meta-schema, denoted as $T_G=(\mathcal{L},\mathcal{R})$, is a directed graph defined over $\mathcal{L}$, the node types in G, and $\mathcal{R}$, the edge types in G.
\end{define}
For example, Fig.~\ref{meta}(a) shows the meta-schema of the DBLP dataset, which is a directed graph that summarizes all the node types and edge types in the dataset.

\begin{define}\textbf{Meta-Path}~\cite{shi2016survey}
Given a meta-schema $T_G=(\mathcal{L},\mathcal{R})$, a meta-path $\mathcal{P}$, denoted as $L_1\xrightarrow[]{R_1}L_2\xrightarrow[]{R_2}...\xrightarrow[]{R_{k-1}}L_{k}$, is a path on $T_G$ that defines a composite relation $R=R_1\circ R_2\circ...\circ R_{k-1}$ between node type $L_1$ and $L_k$. Here $\circ$ is the relation composition operator. 
\end{define}
We say a path $p = v_1-v_2- ...-v_k$ between $v_1$ and $v_k$ in the network $G$ follows the meta-path $\mathcal{P}$, if $\forall i$, $v_i$ is of type $L_i$. 
We call $p$ as a \textit{meta-path instance} of $\mathcal{P}$.
For example, $P_1$ in Fig.~\ref{meta}(a) is a meta-path defined on the meta-schema of the DBLP dataset. 
$P_1$ defines the relation between author nodes. $a_1-p_1-c_1-p_2-a_2$ is a meta-path instance of $P_1$, indicating that author $a_1$ and $a_2$ are similar, because they publish their works, $p_1$ and $p_2$, in the same conference $c_1$.

\begin{define}{\textbf{Meta-Graph}~\cite{zhao2017meta}.}
Given a meta-schema $T_G=(\mathcal{L},\mathcal{R})$, a meta-graph $\mathcal{M}$, is a directed acyclic graph (DAG) with a single source node and a single sink node.
The nodes and edges in $\mathcal{M}$ are confined to $\mathcal{L}$ and $\mathcal{R}$, respectively.
\end{define}
We define the \textit{meta-graph instance} in the same manner as to how we define the meta-path instance. 
For example, $P_4$ in Fig. \ref{meta}(a) is a meta-graph defined on the meta-schema of the DBLP dataset. 
Similar to meta-path $P_1$, $P_4$ also defines the relation between the author nodes. 
However, $P_4$ conveys richer and more complex semantics as compared to $P_1$.
$a_1-p_1-c_1(t_1)-p_2-a_2$ is a meta-graph instance of $P_4$. 
It indicates that author $a_1$ and $a_2$ are similar, because they publish their works, $p_1$ and $p_2$, in the same conference $c_1$, and furthermore, $p_1$ and $p_2$ share the same keyword $t_1$.

In this study, we sometimes refer to meta-paths and meta-graphs collectively as \textit{semantic structures}.
For brevity, we denote a semantic structure using its component node types. 
For example, we denote $P_1$ in Fig.~\ref{meta}(a) as $APCPA$, and $P_3$ as $APA(C)PA$.
All semantic structures used in this study begin and end with the same node type, i.e., the target type. 
For example, as shown in Fig.~\ref{meta}(a), the target type of the DBLP dataset is the author (A).
\section{The Proposed Model}\label{sec:model}
In this section, we propose a novel Higher-order Attribute-Enhancing Graph Neural Network (HAE$_{GNN}$). 
In particular, we elaborate its three main components, i.e., Semantic-Based Convolutional Layer (SCL), Content-Based Self-Attention Layer (CAL), and Higher-order Attribute-Enhancing Framework (HAE).

We present the overall framework of HAE$_{GNN}$ in Fig.~\ref{framework}. 
As shown in Fig.~\ref{framework}(c), HAE$_{GNN}$ models contain one SCL followed by a number of CALs. 
The output node embedding of each component layer serves as the input of the next layer.
Each layer enhances the node features with either semantics, in the case of SCL, or content-based nodes' interactions, in the case of CAL.
We refer to the total number of SCL/CAL layers as the \textit{order} of a HAE$_{GNN}$. 
We denote HAE$_{GNN}$ models with order equals to 2 (i.e., are composed of one SCL and one CAL) as HAE$_{GNN}^{2l}$, and denote those with order equals to 3 (i.e., are composed of one SCL followed by two CALs) as HAE$_{GNN}^{3l}$, and so forth.

\subsection{Semantic-Based Convolutional Layer}
SCL enhances the input node features with rich semantics contained in multiple meta-paths and meta-graphs. 
Fig.~\ref{framework}(a) shows the structure of SCL. 
Specifically, for each pair of target nodes, SCL measures a multi-semantic structure based similarity between those two component nodes.
The similarity is parameterized by the weights of all semantic structures. 
In this way, SCL constructs an adjacent matrix, denoted as $A$ in Fig.~\ref{framework}(a), that contains the similarities between all target nodes.
After that, SCL leverages convolutional structures to fuse the embeddings of the target nodes based on their similarities. 
As a result, SCL learns semantics in the sense that semantically close nodes also have similar embeddings.

Before introducing the multi-semantic structure-based node similarity, we first show how to compute the similarity between two nodes using a single meta-path.
Commuting matrices~\cite{shi2016survey} can be leveraged for such a purpose.
Given a meta-path $\mathcal{P}=L_1L_2...L_k$, its corresponding commuting matrix is defined as $\boldsymbol{C}_{\mathcal{P}}= \boldsymbol{W}_{L_1L_2}\cdot\boldsymbol{W}_{L_2L_3}\cdot ...\cdot\boldsymbol{W}_{L_{k-1}L_k}$, where $\boldsymbol{W}_{L_iL_j}$ is the adjacency matrix between type $L_i$ nodes and type $L_j$ nodes. 
For example, for $P_1 = APCPA$ in Fig.~\ref{meta}(a), its commuting matrix can be computed as $\boldsymbol{C}_{P_1}=\boldsymbol{W}_{AP}\cdot\boldsymbol{W}_{PC}\cdot \boldsymbol{W}_{CP}\cdot\boldsymbol{W}_{PA}= \boldsymbol{W}_{AP}\cdot\boldsymbol{W}_{PC}\cdot \boldsymbol{W}^{\mathrm{T}}_{PC} \cdot \boldsymbol{W}^{\mathrm{T}}_{AP}$, where the superscript $\mathrm{T}$ denotes matrix transposition. 
Note that given $\boldsymbol{C}_\mathcal{P}$, each of its element $\boldsymbol{C}_\mathcal{P}(i,j)$ represents the count of meta-path instances between node $v_{1i}\in L_1$ and node $v_{kj}\in L_k$ under meta-path $\mathcal{P}$. 
Intuitively, large meta-path instances count reflects close linkage, therefore, $\boldsymbol{C}_\mathcal{P}(i,j)$ reveals the similarity between $v_{1i}$ and $v_{kj}$.

Using meta-graphs to measure the nodes' similarity is more complicated because meta-graphs have complex structures. 
Similar to the case of meta-paths, we leverage commuting matrices. 
Given a meta-graph, we adopt the approach in~\cite{zhao2017meta} to compute its commuting matrix. 
Specifically, we treat a meta-graph as a synthesis of multiple meta-paths. 
As discussed above, for each component meta-path, we can represent its commuting matrix as a sequence of multiplications between adjacency matrices. 
We then merge the multiplication sequences of all the component meta-paths to get the commuting matrix of the meta-graph, and we leverage Hadamard Product, i.e., element-wise product, for a combination when the sub-sequences differ. 
For example, meta-graph $P_3 = APA(C)PA$ in Fig.~\ref{meta}(a) can be split into two meta-paths, i.e., $(A,P,A,P,A)$ and $(A,P,C,P,A)$. 
The commuting matrix of the former can be calculated by $\boldsymbol{W}_{AP}\cdot\boldsymbol{W}_{PA}\cdot \boldsymbol{W}_{AP}\cdot\boldsymbol{W}_{PA}$, and the later by $\boldsymbol{W}_{AP}\cdot\boldsymbol{W}_{PC}\cdot \boldsymbol{W}_{CP}\cdot\boldsymbol{W}_{PA}$. 
Then, we can compute the commuting matrix of  $P_3$ by $\boldsymbol{C}_{P_3}= \boldsymbol{W}_{AP}\cdot((\boldsymbol{W}_{PA}\cdot \boldsymbol{W}_{AP})\odot(\boldsymbol{W}_{PC}\cdot \boldsymbol{W}_{CP}))\cdot\boldsymbol{W}_{PA}$. 
Note how the distinct parts, $\boldsymbol{W}_{PA}\cdot \boldsymbol{W}_{AP}$ and $\boldsymbol{W}_{PC}\cdot \boldsymbol{W}_{CP}$, are combined using Hadamard Product.

Inspired by how \cite{peng2019fine} fuses multiple meta-paths for an instance based social event similarity, we then define a semantic structure instance based node similarity (SemSim). 
SemSim simultaneously leverages multiple meta-paths and meta-graphs to measure the similarity between two target nodes. 
We present the definition of SemSim as follows.
\begin{define}{\textbf{Semantic Structure Instances Based Node Similarity (SemSim).}}\label{def:SemSim}
Given a collection of meta-paths and meta-graphs $\boldsymbol{P} = {\{P_m\}}_{m=1}^M$ that start and end with the target type $L_t$, the SemSim between two nodes $v_{ti}\in L_t$ and $v_{tj}\in L_t$ is defined as:
\begin{equation}
SemSim(v_{ti},v_{tj}) = \sum_{m=1}^M\omega_m\frac{2\times\boldsymbol{C}_{P_m}(i,j)}{\boldsymbol{C}_{P_m}(i,i)+ \boldsymbol{C}_{P_m}(j,j)},
\end{equation}
\end{define}
\noindent where $\omega_m$ denotes the weight of $P_m$ and we have a trainable parameter vector $\boldsymbol{\omega} = [\omega_1,\omega_2,...,\omega_M]$ that contains the weights of all meta-paths and meta-graphs. 
$\boldsymbol{\omega}$ is the weights vector shown in Fig.~\ref{framework}(a).
SemSim is asymmetric, i.e., $SemSim(v_{ti},v_{tj})\neq SemSim(v_{tj},v_{ti})$. 
SemSim is self-maximum, i.e., $SemSim(v_{ti},v_{tj}) \in [0,1]$, and $SemSim(v_{ti},v_{ti}) = SemSim(v_{tj},v_{tj}) = 1$.
$SemSim(v_{ti},v_{tj})$ essentially calculates a weighted summation of normalized meta-path/meta-graph instance counts. 
Note that in the fractional term $\frac{2\times\boldsymbol{C}_{P_m}(i,j)}{\boldsymbol{C}_{P_m}(i,i)+ \boldsymbol{C}_{P_m}(j,j)}$, the numerator counts the instances between the two nodes under $P_m$, while the denominator normalizes the term with the instances counts between the nodes themselves under $P_m$.

Next, we capture the semantic-based interactions between nodes with convolutional structures. 
Given the target type $L_t$, we construct a semantic-based adjacent matrix $\boldsymbol{A} \in \mathbb{R}^{N\times N}$, where $N$ is the total number of nodes of type $L_t$, and $\boldsymbol{A}_{ij} = SemSim(v_{ti},v_{tj})$.
$\boldsymbol{A}$ is the adjacent matrix shown in Fig.~\ref{framework}(a).
$\boldsymbol{A}$ contains self-connections, and $\boldsymbol{A}_{ii} = SemSim(v_{ti},v_{ti}) = 1$. 
Note that $\boldsymbol{A}$ fuses all semantic structures. 
Therefore, as opposed to \cite{wang2019heterogeneous}, no semantic-level aggregation is needed in this study, and the learned embeddings naturally fall into the same vector space. 
We use the bag-of-words (BOW) representations of the target nodes as their initial features, denoted as $\boldsymbol{X} \in \mathbb{R}^{N\times d}$, where $d$ is the dimension of initial node features.
After that, we apply a multi-layer GCN architecture extending~\cite{kipf2016semi} to heterogeneous networks. 
Note that our approach is different from \cite{kipf2016semi}. 
The adjacency matrix in \cite{kipf2016semi} is directly constructed from the raw network, assuming that the network is homogeneous. 
In contrast, $\boldsymbol{A}$ in this study incorporates rich heterogeneous semantic information that are contained in meta-paths as well as meta-graphs.
The proposed GCN architecture with sub-layers following the propagation rules:
\begin{equation}
\boldsymbol{H}^{(l+1)} = \sigma(\boldsymbol{D}^{-\frac{1}{2}}\boldsymbol{A}\boldsymbol{D}^{-\frac{1}{2}}\boldsymbol{H}^{(l)}\boldsymbol{W}^{(l)}),
\end{equation}
\noindent where $\boldsymbol{D}$ is a diagonal matrix with $\boldsymbol{D}_{ii} = \sum_j \boldsymbol{A}_{ij}$, and $\boldsymbol{W}^{(l)}$ is a layer-specific trainable weight matrix. 
$\sigma(\cdot)$ is an activation function such as $ReLU$ or $Sigmoid$. $\boldsymbol{H}^{(0)} = \boldsymbol{X}$ is the input node features, and $\boldsymbol{H}^{(l)} \in \mathbb{R}^{N\times d}$ the output node features of the $l^{th}$ layer.

\subsection{Content-Based Self-Attention Layer}
CAL enhances the input node features with the content-based interactions between nodes. 
As shown in Fig.~\ref{framework}(b), CAL first constructs a multi-semantic structure based adjacency matrix, denoted as $\boldsymbol{A}^{\prime}$ in Fig.~\ref{framework}(b), then leverages self-attention~\cite{bahdanau2014neural} to compute the representation of each node by paying attention to its neighbors.

First, given a collection of meta-paths and meta-graphs that both start and end with the target node type $L_t$, we can construct an adjacency matrix $\boldsymbol{A}^{\prime}\in\mathbb{R}^{N\times N}$, where $N$ stands for the total number of nodes of type $L_t$. 
$\boldsymbol{A}^{\prime}$ is a binary matrix, and $\boldsymbol{A}^{\prime}_{ij} = 1$ only if there exist at least 1 meta-path/meta-graph instance between node $v_{ti}$ and node $v_{tj}$. 
For each node $v_{ti} \in L_t$, $\boldsymbol{A}^{\prime}$ defines its \textit{first-order semantic structures based neighborhood} as $\mathcal{N}_i = \{v_{tj}|\boldsymbol{A}^{\prime}_{ij} = 1\}$. 
Note that $v_{ti}$ is also in its neighborhood.

Next, we devise an attention mechanism to perform self-attention on the target nodes.
Note that unlike \cite{velivckovic2017graph}, which pays attention to all immediate neighboring nodes in the raw network, we perform a masked attention that pays attention to the semantic structures based neighborhoods as defined by $\boldsymbol{A}^{\prime}$. 
The attention mechanism is denoted as $a:\mathbb{R}^{d^{\prime}\times d^{\prime}}\rightarrow \mathbb{R}$, where $d^{\prime}$ is the output dimension of CAL. 
$a$ is a single feed-forward layer with non-linearity. 
$a$ takes the linearly transformed representations of two nodes as input, and output an \textit{attention coefficient}:
\begin{equation}
e_{ij} = a(\boldsymbol{W}\boldsymbol{v}_{ti},\boldsymbol{W}\boldsymbol{v}_{tj}) = \sigma({\boldsymbol{a}}^{\mathrm{T}}[\boldsymbol{W}\boldsymbol{v}_{ti}||\boldsymbol{W}\boldsymbol{v}_{tj}]).
\end{equation}
\noindent Here, $\boldsymbol{v}_{ti}\in\mathbb{R}^{d}$ and $\boldsymbol{v}_{tj}\in\mathbb{R}^{d}$ stand for the input representations of node $v_{ti}$ and node $v_{tj}$, respectively. 
$\boldsymbol{W} \in \mathbb{R}^{d^{\prime}\times d}$ is a weight matrix that transforms the input node features to higher-level features. 
$\boldsymbol{a} \in \mathbb{R}^{2d^{\prime}}$ is a weight vector. 
$\sigma(\cdot)$ denotes the nonlinear function, and $||$ stands for the concatenation operation. 
Note that $\boldsymbol{W}$ and $\boldsymbol{a}$ are shared among all node pairs.
The attention coefficient $e_{ij}$ indicates the importance of $v_{tj}$’s representation to $v_{ti}$.

For the attention coefficients to be comparable across different nodes, we normalize them across all neighboring nodes. 
We leverage softmax function and get the \textit{normalized attention coefficient}:
\begin{equation}
\alpha_{ij} = \text{softmax}(e_{ij}) = \frac{\text{exp}(\sigma(\boldsymbol{a}^T[\boldsymbol{W}\boldsymbol{v}_{ti}||\boldsymbol{W}\boldsymbol{v}_{tj}]))}{\sum_{n \in \mathcal{N}_i}\text{exp}(\sigma(\boldsymbol{a}^T[\boldsymbol{W}\boldsymbol{v}_{ti}||\boldsymbol{W}\boldsymbol{v}_{tn}]))},
\end{equation}
\noindent where $\mathcal{N}_i$ is a set of $v_{ti}$'s first-order semantic structure-based neighbors according to $\boldsymbol{A}^{\prime}$. 
Note that $\alpha_{ij}$ is asymmetric.
The output representation of $v_{ti}$ can then be computed by paying attention to its neighbors using the normalized attention coefficients:
\begin{equation}
{\boldsymbol{v}^{\prime}_{ti}} = \sigma\Bigg(\sum_{{v}_{tj} \in \mathcal{N}_i}\alpha_{ij}\boldsymbol{W}\boldsymbol{v}_{tj}\Bigg).
\end{equation}

We further employ multi-head attention~\cite{vaswani2017attention} to stabilize the learning process. 
Specifically, we train $H$ independent attention mechanisms, and concatenate their outputs as the final representation:
\begin{equation}
{\boldsymbol{v}^{\prime}_{ti}} = {\parallel}^{H}_{h=1}\sigma\Bigg(\sum_{{v}_{tj} \in \mathcal{N}_i}\alpha^{h}_{ij}\boldsymbol{W}^{h}\boldsymbol{v}_{tj}\Bigg).
\end{equation}
\noindent Here, $\alpha^{h}_{ij}$ stands for the head-wise normalized attention coefficients, and $\boldsymbol{W}^{h}$ stands for the head-wise linear transformation matrix.
In this study, we set the output dimension of each head to $d^{\prime} = d/H$, such that the output dimension of CAL is equal to its input dimension.

\subsection{Higher-Order Attribute-Enhancing Framework}
As shown in Fig.~\ref{framework}(c), the HAE framework stacks multiple SCLs and CALs together. 
It constructs a higher-order attribute-enhancing architecture that enhances the input node features with both semantics (through SCLs) and content-based nodes' interactions (through CALs) in a layer-by-layer manner.

\begin{figure}[t]
    \centering
    \includegraphics[width =9cm]{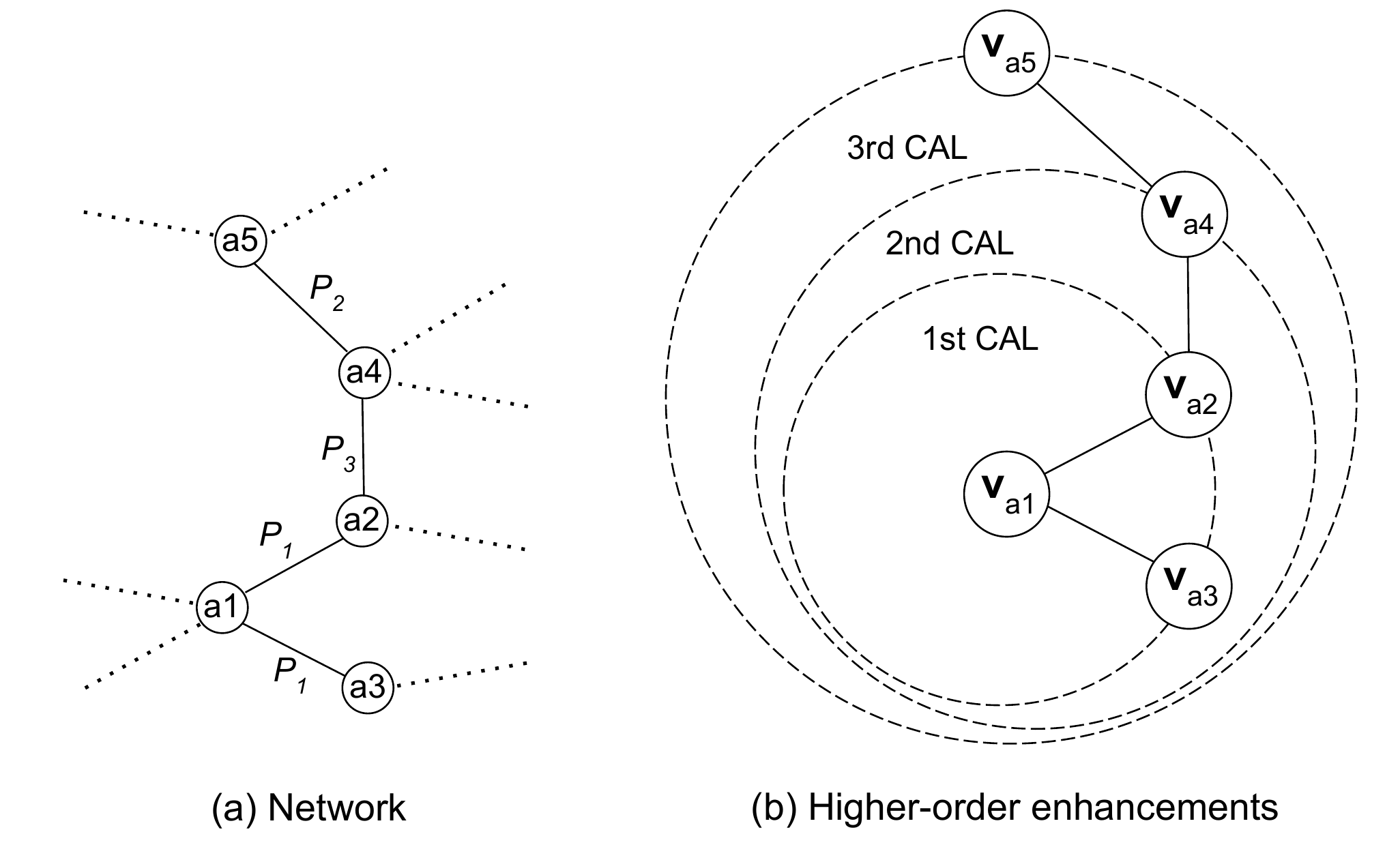}
    \caption{An illustration of how HAE$_{GNN}^{4l}$ captures the content-based nodes' interactions in DBLP. (a) shows part of the network mapped by $\boldsymbol{A}^{\prime}$, where $a_1$ through $a_5$ are authors, and $P_1$ through $P_3$ are semantic structures listed in Fig. \ref{meta}. (b) shows how the three CALs in HAE$_{GNN}^{4l}$ gradually enhance the author $a_1$'s embedding with the embeddings of its higher-order neighbors. $\boldsymbol{v}_{a1}$ through $\boldsymbol{v}_{a5}$ are authors' embeddings that have been enhanced by the SCL in HAE$_{GNN}^{4l}$ (and yet to be enhanced by the CALs). }
    \label{higher_order_enhance}
\end{figure}

We take HAE$_{GNN}$ with order equals to 4 (denoted as HAE$_{GNN}^{4l}$) as a concrete example to show how the higher-order architecture works. 
As shown in Fig.~\ref{framework}(c), HAE$_{GNN}^{4l}$ contains four component layers, i.e., one SCL followed by three CALs. 
The SCL enhances the initial BOW node features with rich heterogeneous semantics contained in meta-paths and meta-graphs. 
After then, by stacking the three CALs, HAE$_{GNN}^{4l}$ is able to capture the content-based interaction between the first-order as well as the \textit{higher-order semantic structures based neighborhoods}. 
Fig.~\ref{higher_order_enhance} illustrates this process using the DBLP dataset as an example. 
A CAL enhances each author's embedding with the embeddings of its first-order semantic structures based neighbors. 
As a result, the first CAL enhances $\boldsymbol{v}_{a1}$ with $\boldsymbol{v}_{a2}$ and $\boldsymbol{v}_{a3}$, and enhances $\boldsymbol{v}_{a2}$ with $\boldsymbol{v}_{a1}$ and $\boldsymbol{v}_{a4}$. 
The second CAL then repeats this process and enhances $\boldsymbol{v}_{a1}$ with $\boldsymbol{v}_{a2}$. 
At this point, $\boldsymbol{v}_{a2}$ has been enhanced with $\boldsymbol{v}_{a4}$ by the first CAL, and therefore, the second CAL indirectly enhances $\boldsymbol{v}_{a1}$ with $\boldsymbol{v}_{a4}$. 
Similarly, the third CAL enhances $\boldsymbol{v}_{a1}$ with $\boldsymbol{v}_{a5}$. 
In this way, HAE$_{GNN}^{4l}$ incorporates author $a_1$'s embedding with the embeddings of its first-order (author $a_2$ and $a_3$), second-order (author $a_4$), and third-order (author $a_5$) neighbors.

As shown in Fig. \ref{framework}(c), HAE$_{GNN}$ models take a matrix that contains the initial features of the target type nodes as input, and output the enhanced nodes' embeddings. 
The enhancing process is guided by task-specific loss functions. 
Taking the node classification task as an example, during training, HAE$_{GNN}$ models, as semi-supervised models, observe the labels of a subset of the target nodes, and minimize the cross-entropy between the ground-truths and the predictions. 
The loss function can be formalized as:
\begin{equation}
L=-\sum_{v\in \mathcal{V}_t}\boldsymbol{y}_v \text{log}(\text{softmax}(\boldsymbol{W}\boldsymbol{v} + \boldsymbol{b})).
\end{equation}
\noindent Here, $\mathcal{V}_t$ is a subset of target nodes with pre-known labels.
$\boldsymbol{y}_v \in \mathbb{R}^{c}$ is a one-hot vector that represents the label of node $v$, where $c$ is the number of distinct classes. 
$\boldsymbol{v} \in \mathbb{R}^{d}$ is the representation of $v$ output by HAE$_{GNN}$. 
$\boldsymbol{W} \in \mathbb{R}^{c\times d}$ and $\boldsymbol{b} \in \mathbb{R}^{c}$ are the parameters of the fully connected layer that performs linear transformation.

Note the HAE framework allows stacking of an arbitrary number of SCLs and CALs in an arbitrary sequence.
Therefore, under the HAE framework, there are other variations besides HAE$_{GNN}$, such as models that contain solely SCLs (denoted as HAE$_{SCL}$) and models that contain solely CALs (denoted as HAE$_{CAL}$). 
Such flexibility provides users the freedom to construct task-oriented models.
For heterogeneous network analysis tasks including node classification, node clustering, and visualization, we adopt the HAE$_{GNN}$ architecture as elaborated above. 
To verify the superiority of the HAE$_{GNN}$ architecture, in Section \ref{sec:exper}, we experiment on HAE$_{GNN}$ of different orders, as well as other variations under the HAE framework for comparison.
\section{Experiments}\label{sec:exper}

We evaluate the proposed HAE$_{GNN}$ framework through graph analysis tasks, including node classification, node clustering, and visualization. 
All these are classic tasks that are commonly performed by network representation learning studies \cite{cui2018survey}. 
Furthermore, we show the merits of our semantic enhancing approach and higher-order attribute-enhancing strategy with experimental results, memory consumption and parameter analysis.

\subsection{Datasets}
We conduct experiments on three datasets. 
We list out the meta-schemas along with the semantic structures used in the experiments for all datasets in Fig.~\ref{meta}. 

\textbf{DBLP:}
We adopt a subset of DBLP as extracted by \cite{wang2019heterogeneous}. 
This subset contains 27,194 nodes, including 14,328 papers (P), 4,057 authors (A), 20 conferences (C) and 8,789 terms (T). 
The subset contains 122,393 edges, with detailed statistics in \cite{wang2019heterogeneous}. 
The authors are labeled by their research areas, and there are four classes, i.e., Database, Data Mining, AI and Information Retrieval.
We use the BOW representations of keywords (terms) as the initial features of the author nodes. 

\textbf{IMDB:} 
We extract a subset of IMDB which contains 9,692 nodes, including 3,627 movies (M), 4,340 actors (A), 1,714 directors (D) and 11 rating groups (R). 
The dataset contains 18,132 edges, including 10,878 between M and A, 3,627 between M and D, and 3,627 between M and R. 
The movies are labeled by their genres, and there are three classes, i.e., Action, Comedy, and Drama. 
We use the BOW representations of plots as the initial features of the movies.

\textbf{HUAWEI:}
A dataset provided by DIGIX Huawei Global Smartphone Theme Design Competition.
The dataset contains de-identified users' demographics, device usage behavior, and app usage behavior. 
There are 10,371 nodes, including 4,200 users (U), 6,131 applications (A) and 40 application types (T). 
There are 248,479 edges, including 8,112 between A and U, 175,381 between A and U, 55,858 between U and T, and 9,128 between U and U (an edge between two users indicates that they follow each other in social media apps such as WeChat). 
We label the users according to their age groups, and there are six groups, including $age<18$, $19\sim26$, $27\sim35$, $36\sim44$, $45\sim53$, and $>54$. 
We use the BOW representations of behavioral attributes as the users' initial features.

\subsection{Baselines}
To verify the effectiveness of the HAE$_{GNN}$ framework, we experiment on HAE$_{GNN}$ with 1 SCL and 3 CALs, namely HAE$_{GNN}^{4l}$, to represent the performances of HAE$_{GNN}$.
We compare with random walk or attributed network based unsupervised baselines as well as GNN based semi-supervised baselines. 
The baselines are as follows:

\textbf{Unsupervised methods.} 
We use DeepWalk \cite{perozzi2014deepwalk} (Deepw.), Metapath2vec~\cite{dong2017metapath2vec} (Metap.), DANE~\cite{gao2018deep} and \FIXED{HONE~\cite{RossiHigherOrder2018}} as unsupervised baselines.
DANE is an attributed homogeneous network.
\FIXED{HONE is a higher-order network representation learning using motifs for homogeneous networks.}
For experiments on DeepWalk, DANE and \FIXED{HONE}, we ignore the heterogeneity of the graph data.
For metapath2vec, we experiment on all meta-paths and report the best results.

\textbf{Semi-supervised methods.} 
We use GNN based models, including SDNE~\cite{wang2016structural}, GCN~\cite{kipf2016semi}, GAT~\cite{velivckovic2017graph}, HAN~\cite{wang2019heterogeneous}, GTN~\cite{yun2019graph}, RSHN~\cite{zhu2019relation}, HetGNN~\cite{zhang2019heterogeneous} (HetG.), HGT~\cite{hu2020heterogeneous} and MAGNN~\cite{fu2020magnn} (MAG.) as semi-supervised baselines.
For SDNE, GCN and GAT, we leverage a single meta-path to transform the datasets into homogeneous graphs before conducting the experiments. 
We experiment on all meta-paths and report the best results.
HAN, GTN, RSHN, HetGNN, HGT and MAGNN are typical heterogeneous graph neural network models.

In addition, we also compare HAE$_{GNN}^{4l}$ with other variants under the HAE framework. 
We experiment with three different architectures under the HAE framework, i.e., HAE$_{GNN}$ (contains an SCL followed by a number of CALs), HAE$_{SCL}$ (contains SCLs solely) and HAE$_{CAL}$ (contains CALs solely). 
For HAE$_{GNN}$, besides HAE$_{GNN}^{4l}$, which contains one SCL and three CALs, we also experiment on HAE$_{GNN}^{2l}$, which contains one SCL and one CAL. 
For HAE$_{SCL}$ and HAE$_{CAL}$, we experiment on HAE$_{SCL}^{2l}$, which contains two SCLs, and HAE$_{CAL}^{4l}$, which contains four CALs, respectively.

\subsection{Experimental Settings}
For the HAE models, we set the learning rate to 0.0003, the number of sub-layers in SCLs to 2, and the number of attention heads in CALs to 8.
We set the dropout rate of the first CAL to 0.4, and for each of the following CALs, we use one-half of the previous CAL's dropout rate.
The $\boldsymbol{\omega}$ in each SCL, which contains the weights of the semantic structures, is initialized with Xavier to guarantee that the weights are normalized and sum to 1.
We optimize our models with Adam~\cite{kingma2015adam}.
For random walk based models, i.e., DeepWalk and Metapath2vec, we set the window size to 10, the walk length to 40, and the number of walks per node to 80.
For a fair comparison, we set the embedding dimension to 64 for all models, and use the training ratio from 0.2 to 0.8 for all semi-supervised models.
For the semi-supervised baseline methods, their parameters are set following the original literature.
We conduct all experiments on a 64 core Intel Xeon CPU E5-2680 v4@2.40GHz with 512GB RAM and 8$\times$NVIDIA Tesla P100-PICE GPUs.
For the acceleration of all CPUs calculations, we have adopted a parallelized processing of binding 64 cores.

\renewcommand{\arraystretch}{1.3}
\begin{table*}[tp]
\caption{Node classification results (\%) compared to the baselines.}\label{classfication_vs_base}  
\centering
\scriptsize
\scalebox{1}{
\begin{tabular}{c|cc|cccc|ccc|cccccc|c}
\hlinewd{0.7pt}
\multicolumn{1}{c} { } &  &  & Deepw. & Metap. & DANE & \FIXED{HONE} & SDNE & GCN & GAT & GTN & HAN & RSHN & MAG. & HGT & HetG. & HAE$_{GNN}^{4l}$\\ 
\hline
\multirow{8}{*}{\textbf{\rotatebox{90}{DBLP}} } & \multirow{4}{*}{\textbf{\rotatebox{90}{Ma-F1}}}
                           & \textbf{20\%} & 73.95 & 81.05 & 80.99 & \FIXED{77.05} & 81.41 & 83.95 & 84.75 & 83.71 & 89.44 & 88.01 & 89.28 & 90.90 & 90.85 & \textbf{91.54}\FIXED{$\pm$.09}  \\
                        &  & \textbf{40\%} & 77.71 & 82.73 & 81.43 & \FIXED{80.43} & 83.03 & 84.88 & 85.42 & 85.36 & 90.74 & 90.28 & 90.83 & 91.11 & 91.77 & \textbf{92.91}\FIXED{$\pm$.06}  \\
                        &  & \textbf{60\%} & 79.26 & 83.88 & 83.95 & \FIXED{83.32} & 85.00 & 85.03 & 86.86 & 87.84 & 91.02 & 91.00 & 91.74 & 92.55 & 92.91 & \textbf{93.07}\FIXED{$\pm$.05}  \\
                        &  & \textbf{80\%} & 80.52 & 84.21 & 85.01 & \FIXED{84.06} & 85.22 & 85.37 & 87.27 & 90.51 & 91.37 & 91.91 & 92.05 & 93.04 & 93.15 & \textbf{93.69}\FIXED{$\pm$.03} \\
                        \cline{2-17}
                        & \multirow{4}{*}{\textbf{\rotatebox{90}{Mi-F1}}}
                           & \textbf{20\%} & 73.58 & 81.68 & 81.60 & \FIXED{76.63} & 82.54 & 83.00 & 84.92 & 84.27 & 89.16 & 87.81 & 90.70 & 91.51 & 90.70 & \textbf{91.64}\FIXED{$\pm$.10}  \\
                        &  & \textbf{40\%} & 77.27 & 82.59 & 82.58 & \FIXED{81.55} & 83.19 & 83.51 & 86.11 & 86.72 & 90.25 & 90.04 & 91.31 & 92.28 & 91.95 & \textbf{92.40}\FIXED{$\pm$.08}  \\
                        &  & \textbf{60\%} & 80.44 & 83.05 & 82.95 & \FIXED{83.61} & 84.06 & 84.73 & 87.73 & 88.04 & 91.15 & 91.05 & 92.05 & 93.09 & 93.15 & \textbf{93.17}\FIXED{$\pm$.05}  \\
                        &  & \textbf{80\%} & 81.93 & 83.21 & 83.66 & \FIXED{84.13} & 84.41 & 84.80 & 88.26 & 91.35 & 91.77 & 92.06 & 92.37 & 93.25 & 93.77 & \textbf{93.82}\FIXED{$\pm$.02} \\
                          \hline
\multirow{8}{*}{\textbf{\rotatebox{90}{IMDB}}} & \multirow{4}{*}{\textbf{\rotatebox{90}{Ma-F1}}} 
                           & \textbf{20\%} & 35.03 & 42.17 & 42.04 & \FIXED{38.88} & 45.38 & 45.72 & 51.53 & 51.37 & 52.09 & 51.94 & 52.08 & 52.52 & 53.74 & \textbf{54.60}\FIXED{$\pm$.10}  \\
                        &  & \textbf{40\%} & 37.18 & 43.25 & 42.87 & \FIXED{41.09} & 45.95 & 47.83 & 51.99 & 53.58 & 53.39 & 53.22 & 54.05 & 54.79 & 55.01 & \textbf{56.69}\FIXED{$\pm$.05}  \\
                        &  & \textbf{60\%} & 39.57 & 44.36 & 44.02 & \FIXED{43.43} & 48.66 & 49.19 & 52.80 & 55.16 & 54.62 & 54.49 & 56.28 & 57.00 & 58.77 & \textbf{60.77}\FIXED{$\pm$.08}  \\
                        &  & \textbf{80\%} & 40.21 & 45.93 & 46.69 & \FIXED{44.40} & 49.77 & 50.51 & 53.46 & 57.63 & 56.88 & 56.99 & 57.59 & 58.64 & 60.01 & \textbf{62.96}\FIXED{$\pm$.05} \\
                        \cline{2-17}
                        & \multirow{4}{*}{\textbf{\rotatebox{90}{Mi-F1}}}
                           & \textbf{20\%} & 35.62 & 43.40 & 43.06 & \FIXED{38.90} & 45.35 & 45.91 & 51.72 & 51.91 & 52.05 & 51.90 & 53.06 & 53.75 & 53.09 & \textbf{54.00}\FIXED{$\pm$.07}  \\
                        &  & \textbf{40\%} & 38.75 & 44.27 & 44.15 & \FIXED{41.47} & 48.11 & 48.26 & 52.06 & 52.46 & 53.53 & 53.40 & 54.90 & 55.05 & 55.88 & \textbf{56.73}\FIXED{$\pm$.05}  \\
                        &  & \textbf{60\%} & 40.09 & 45.01 & 45.25 & \FIXED{43.80} & 49.03 & 49.94 & 53.17 & 54.95 & 57.65 & 57.70 & 57.46 & 57.37 & 59.48 & \textbf{60.25}\FIXED{$\pm$.05}  \\
                        &  & \textbf{80\%} & 41.68 & 45.66 & 45.70 & \FIXED{45.01} & 49.49 & 50.78 & 53.96 & 56.82 & 57.17 & 58.07 & 58.91 & 59.48 & 60.61 & \textbf{62.61}\FIXED{$\pm$.04} \\
                          \hline
\multirow{8}{*}{\textbf{\rotatebox{90}{HUAWEI}}} & \multirow{4}{*}{\textbf{\rotatebox{90}{Ma-F1}}}
                           & \textbf{20\%} & 27.53 & 32.59 & 32.85 & \FIXED{30.33} & 33.14 & 33.03 & 33.58 & 33.79 & 37.99 & 37.57 & 38.80 & 38.92 & 39.64 & \textbf{39.81}\FIXED{$\pm$.12}  \\
                        &  & \textbf{40\%} & 29.20 & 33.72 & 33.57 & \FIXED{32.79} & 33.96 & 34.07 & 34.63 & 35.71 & 38.58 & 38.70 & 39.62 & 39.24 & 40.12 & \textbf{40.75}\FIXED{$\pm$.08}  \\
                        &  & \textbf{60\%} & 30.87 & 34.84 & 33.99 & \FIXED{33.62} & 34.48 & 34.95 & 35.85 & 38.42 & 39.13 & 39.91 & 40.15 & 41.38 & 41.03 & \textbf{41.44}\FIXED{$\pm$.07}  \\
                        &  & \textbf{80\%} & 31.24 & 35.90 & 35.68 & \FIXED{34.27} & 36.62 & 36.24 & 37.15 & 40.36 & 40.35 & 40.70 & 41.39 & 42.64 & 42.69 & \textbf{42.83}\FIXED{$\pm$.03} \\
                        \cline{2-17}
                        & \multirow{4}{*}{\textbf{\rotatebox{90}{Mi-F1}}} 
                           & \textbf{20\%} & 28.82 & 32.33 & 32.18 & \FIXED{30.58} & 33.06 & 33.18 & 34.72 & 34.42 & 38.07 & 38.04 & 38.29 & 39.32 & 40.29 & \textbf{40.70}\FIXED{$\pm$.11}  \\
                        &  & \textbf{40\%} & 30.09 & 32.91 & 32.73 & \FIXED{32.94} & 33.77 & 34.75 & 35.93 & 35.61 & 39.59 & 39.22 & 39.10 & 40.53 & 41.82 & \textbf{41.99}\FIXED{$\pm$.07}  \\
                        &  & \textbf{60\%} & 31.11 & 34.28 & 34.49 & \FIXED{33.88} & 35.01 & 35.09 & 36.48 & 38.89 & 40.40 & 40.43 & 40.40 & 41.48 & 42.79 & \textbf{42.89}\FIXED{$\pm$.07}  \\
                        &  & \textbf{80\%} & 32.76 & 35.45 & 35.33 & \FIXED{35.13} & 36.40 & 36.84 & 37.62 & 40.76 & 41.68 & 41.80 & 41.79 & 42.84 & 43.73 & \textbf{43.97}\FIXED{$\pm$.04} \\
                        \hlinewd{0.7pt}                      
\end{tabular}}
\end{table*}

\subsection{Node Classification}\label{sec:classification}
This section evaluates HAE$_{GNN}^{4l}$ by multi-class node classification. 
We adopt a \textit{Logistic Regression} classifier. 
We use $20\%-80\%$ of all targeted nodes for training, and let the classifier observe their labels.
The task is to predict the labels of the remaining nodes. 
\FIXED{We repeat each experiment ten times and take the average results.} 
We report the overall Macro-F1 (Ma-F1) and Micro-F1 (Mi-F1) scores in Table~\ref{classfication_vs_base}.

Overall, the results show that HAE$_{GNN}^{4l}$ outperforms state-of-the-art baselines across all three datasets.
We give more comparative analysis with the representative semi-supervised GNNs.
Compared to GTN, HAE$_{GNN}^{4l}$ improves the Ma-F1 and Mi-F1 scores by up to $7.83\%$ and $7.37\%$ on DBLP, $5.61\%$ and $5.79\%$ on IMDB, and $6.02\%$ and $6.38\%$ on HUAWEI.
This is because HAE$_{GNN}^{4l}$ considers the semantic-based as well as content-based nodes' interactions, while GTN ignores the later for not using self-attentions. 
Compared to HAN, HAE$_{GNN}^{4l}$ improves the Ma-F1 and Mi-F1 scores by at least $2.05\%$ and $2.02\%$ on DBLP, $2.51\%$ and $1.95\%$ on IMDB, and $1.82\%$ and $2.40\%$ on HUAWEI.
This is because HAE$_{GNN}^{4l}$ learns semantics from not only meta-paths but also meta-graphs, and meta-graphs are ignored by HAN. 
Also, HAE$_{GNN}^{4l}$ leverages higher-order architecture to capture the short-range as well as long-range nodes' interdependencies, while the node-level attention architecture in HAN is single-layered and considers only the former. 
Similarly, compared to MAGNN, HAE$_{GNN}^{4l}$ improves the Ma-F1 and Mi-F1 scores by at least $1.33\%$ and $0.94\%$ on DBLP, $2.52\%$ and $0.94\%$ on IMDB, and $1.01\%$ and $2.18\%$ on HUAWEI.
The superiority results show the importance of meta-graph relations in aggregating information.
Compared to RSHN, HAE$_{GNN}^{4l}$ improves the Ma-F1 and Mi-F1 scores by at least $1.78\%$ and $1.76\%$ on DBLP, $2.66\%$ and $2.10\%$ on IMDB, and $1.53\%$ and $2.17\%$ on HUAWEI.
The superiority results show the advantage of multi-hop relationships, such as meta-paths and meta-graphs, than single-hop relations used in the RSHN for information aggregation.
Compared to HGT and HetGNN, HAE$_{GNN}^{4l}$ improves the Ma-F1 and Mi-F1 scores by at least $0.16\%$ and $0.02\%$ on DBLP, $0.86\%$ and $0.25\%$ on IMDB, and $0.14\%$ and $0.10\%$ on HUAWEI. 
Although their performances are comparable, compared to HGT and HetGNN, HAE$_{GNN}^{4l}$ achieves overall surpassing results on DBLP, IMDB and HUAWEI datasets.
The possible reason is that, compared with complex neural network structures, the model with higher-order attribute-enhancing neural network structure is more conducive to representing heterogeneous graphs.

A comparison between the baseline models shows that the GNN based semi-supervised models generally are more effective than the random walk or attributed network based unsupervised models. 
SDNE, GCN, GAT, GTN, HAN, RSHN, HetGNN, HGT, MAGNN and HAE$_{GNN}^{4l}$ largely outperform Metapath2vec, which is the best unsupervised structural heterogeneous network embedding model, in both Ma-F1 and Mi-F1 scores on all datasets (except part of results of DANE and HONE achieve comparable results as Metapath2vec on the three datasets).
Also, models that incorporate heterogeneous semantics perform better than the homogeneous ones, as GTN, HAN, RSHN, MAGNN, HGT and HetGNN give higher Ma-F1 and Mi-F1 scores than SDNE, GCN and GAT on all datasets.
In sum, the node classification results verify the effectiveness of our design, as our models incorporate more abundant semantics, and capture more comprehensive content-based interactions between nodes compared to the baselines.

\renewcommand{\arraystretch}{1.3} 
\begin{table*}[tp]  
  \centering  
  \fontsize{7}{8}\selectfont  
  \caption{Node clustering results (\%) compared to the baselines.}  
  \label{clustering_vs_base}  

    \begin{tabular}{p{2.8cm}<{\centering}|p{1.2cm}<{\centering}p{1.2cm}<{\centering}p{1.2cm}<{\centering}|p{1.2cm}<{\centering}p{1.2cm}<{\centering}p{1.2cm}<{\centering}|p{1.2cm}<{\centering}p{1.2cm}<{\centering}p{1.2cm}<{\centering}}  
    \hlinewd{0.7pt}
    \multirow{2}{*}{Method} &
    
    \multicolumn{3}{c}{\textbf{DBLP}} & \multicolumn{3}{c}{\textbf{IMDB}} &    \multicolumn{3}{c}{\textbf{HUAWEI}} \\ \cline{2-10}
    
    &\textbf{NMI}&\textbf{ARI}&\textbf{FMI}&\textbf{NMI}&\textbf{ARI}&\textbf{FMI}&\textbf{NMI}&\textbf{ARI}&\textbf{FMI}\\
    
     \cline{1-10}
    Deepwalk & 36.32\FIXED{$\pm$.13} & 38.55\FIXED{$\pm$.12} & 48.37\FIXED{$\pm$.13} & 20.87\FIXED{$\pm$.10} & 19.21\FIXED{$\pm$.10} & 29.52\FIXED{$\pm$.09} & 8.82\FIXED{$\pm$.09} & 6.29\FIXED{$\pm$.09} & 18.25\FIXED{$\pm$.10} \cr  
    Metapath2vec & 52.96\FIXED{$\pm$.14} & 56.11\FIXED{$\pm$.15} & 64.18\FIXED{$\pm$.13} & 19.74\FIXED{$\pm$.10} & 21.16\FIXED{$\pm$.11} & 32.80\FIXED{$\pm$.11} & 10.13\FIXED{$\pm$.09} & 8.87\FIXED{$\pm$.09} & 20.59\FIXED{$\pm$.09} \cr
    DANE & 52.78\FIXED{$\pm$.11} & 56.54\FIXED{$\pm$.12} & 63.83\FIXED{$\pm$.10} & 19.65\FIXED{$\pm$.09} & 21.09\FIXED{$\pm$.10} & 32.77\FIXED{$\pm$.09} & 10.27\FIXED{$\pm$.09} & 8.96\FIXED{$\pm$.10} & 21.01\FIXED{$\pm$.10} \cr
    \FIXED{HONE} & \FIXED{45.47}\FIXED{$\pm$.09} & \FIXED{46.30}\FIXED{$\pm$.08} & \FIXED{55.75}\FIXED{$\pm$.10} & \FIXED{21.92}\FIXED{$\pm$.08} & \FIXED{20.78}\FIXED{$\pm$.09} & \FIXED{30.01}\FIXED{$\pm$.10} & \FIXED{11.66}\FIXED{$\pm$.08} & \FIXED{8.80}\FIXED{$\pm$.09} & \FIXED{22.52}\FIXED{$\pm$.07} \cr
    \cline{1-10}
    SDNE & 55.71\FIXED{$\pm$.11} & 58.84\FIXED{$\pm$.12} & 62.26\FIXED{$\pm$.09} & 20.29\FIXED{$\pm$.09} & 22.17\FIXED{$\pm$.10} & 33.33\FIXED{$\pm$.09} & 14.55\FIXED{$\pm$.08} & 11.05\FIXED{$\pm$.07} & 25.11\FIXED{$\pm$.08} \cr
    GCN & 59.71\FIXED{$\pm$.10} & 62.66\FIXED{$\pm$.09} & 66.92\FIXED{$\pm$.08} & 22.67\FIXED{$\pm$.07} & 22.89\FIXED{$\pm$.06} & 32.96\FIXED{$\pm$.06} & 15.24\FIXED{$\pm$.05} & 11.73\FIXED{$\pm$.05} & 25.52\FIXED{$\pm$.05} \cr 
    GAT & 66.23\FIXED{$\pm$.09} & 64.27\FIXED{$\pm$.08} & 74.58\FIXED{$\pm$.09} & 25.63\FIXED{$\pm$.06} & 23.78\FIXED{$\pm$.07} & 32.86\FIXED{$\pm$.06} & 20.71\FIXED{$\pm$.04} & 18.21\FIXED{$\pm$.04} & 31.07\FIXED{$\pm$.05} \cr
     \cline{1-10}
    GTN & 68.97\FIXED{$\pm$.11} & 69.01\FIXED{$\pm$.10} & 80.22\FIXED{$\pm$.11} & 27.59\FIXED{$\pm$.08} & 26.37\FIXED{$\pm$.09} & 38.68\FIXED{$\pm$.09} & 25.82\FIXED{$\pm$.07} & 23.74\FIXED{$\pm$.06} & 34.91\FIXED{$\pm$.06}  \cr
    HAN & 66.51\FIXED{$\pm$.09} & 70.46\FIXED{$\pm$.08} & 77.22\FIXED{$\pm$.09} & 24.95\FIXED{$\pm$.07} & 26.51\FIXED{$\pm$.05} & 34.41\FIXED{$\pm$.06} & 26.93\FIXED{$\pm$.04} & 25.16\FIXED{$\pm$.03} & 33.21\FIXED{$\pm$.04}  \cr
    RSHN & 66.09\FIXED{$\pm$.08} & 69.88\FIXED{$\pm$.06} & 76.40\FIXED{$\pm$.08} & 24.51\FIXED{$\pm$.07} & 26.42\FIXED{$\pm$.06} & 34.15\FIXED{$\pm$.05} & 27.07\FIXED{$\pm$.03} & 25.47\FIXED{$\pm$.04} & 34.73\FIXED{$\pm$.03} \cr
    MAGNN & 69.67\FIXED{$\pm$.06} & 74.68\FIXED{$\pm$.06} & 80.73\FIXED{$\pm$.07} & 25.71\FIXED{$\pm$.06} & 28.81\FIXED{$\pm$.06} & 37.69\FIXED{$\pm$.05} & 27.12\FIXED{$\pm$.03} & 25.89\FIXED{$\pm$.04} & 35.10\FIXED{$\pm$.04}\\
    HGT & 71.99\FIXED{$\pm$.06} & 76.31\FIXED{$\pm$.05} & 83.16\FIXED{$\pm$.06} & 27.09\FIXED{$\pm$.04} & 29.11\FIXED{$\pm$.03} & 40.29\FIXED{$\pm$.04} & 28.00\FIXED{$\pm$.02} & 26.16\FIXED{$\pm$.03} & 36.33\FIXED{$\pm$.02}\\
    HetGNN & 73.00\FIXED{$\pm$.08} & 77.48\FIXED{$\pm$.07} & 84.85\FIXED{$\pm$.07} & 30.86\FIXED{$\pm$.05} & 29.59\FIXED{$\pm$.04} & 42.03\FIXED{$\pm$.04} & 29.25\FIXED{$\pm$.02} & 27.24\FIXED{$\pm$.03} & 37.37\FIXED{$\pm$.03}\\
     \cline{1-10}
    HAE$_{GNN}^{4l}$ & \textbf{77.64}\FIXED{$\pm$.02} & \textbf{79.79}\FIXED{$\pm$.03} & \textbf{85.92}\FIXED{$\pm$.03} & \textbf{32.63}\FIXED{$\pm$.02} & \textbf{31.28}\FIXED{$\pm$.03} & \textbf{44.82}\FIXED{$\pm$.02} & \textbf{30.25}\FIXED{$\pm$.03} & \textbf{28.17}\FIXED{$\pm$.03} & \textbf{38.92}\FIXED{$\pm$.02}\cr
    \bottomrule  
    \end{tabular}  
\end{table*}

\begin{figure*}
\centering
\begin{minipage}{.24\linewidth}
\subfloat[Deepwalk]{\label{visual:a}\includegraphics[width=4cm]{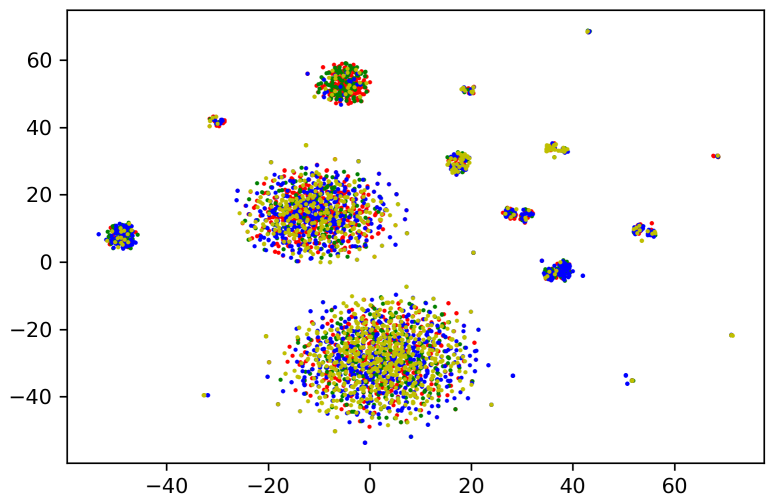}}
\end{minipage}
\begin{minipage}{.24\linewidth}
\subfloat[Metapath2vec]{\label{visual:b}\includegraphics[width=4cm]{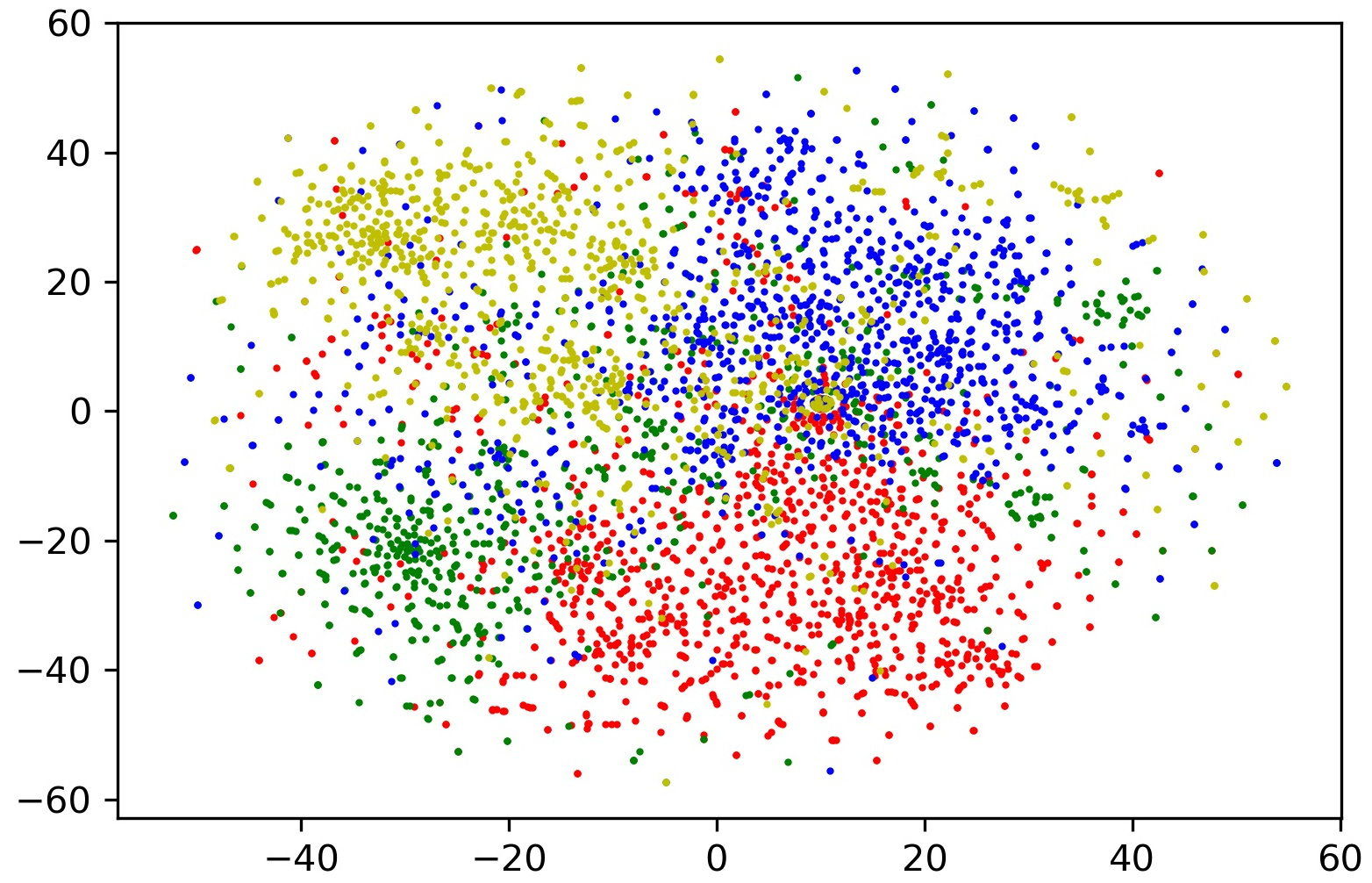}}
\end{minipage}
\begin{minipage}{.24\linewidth}
\subfloat[GCN]{\label{visual:c}\includegraphics[width=4cm]{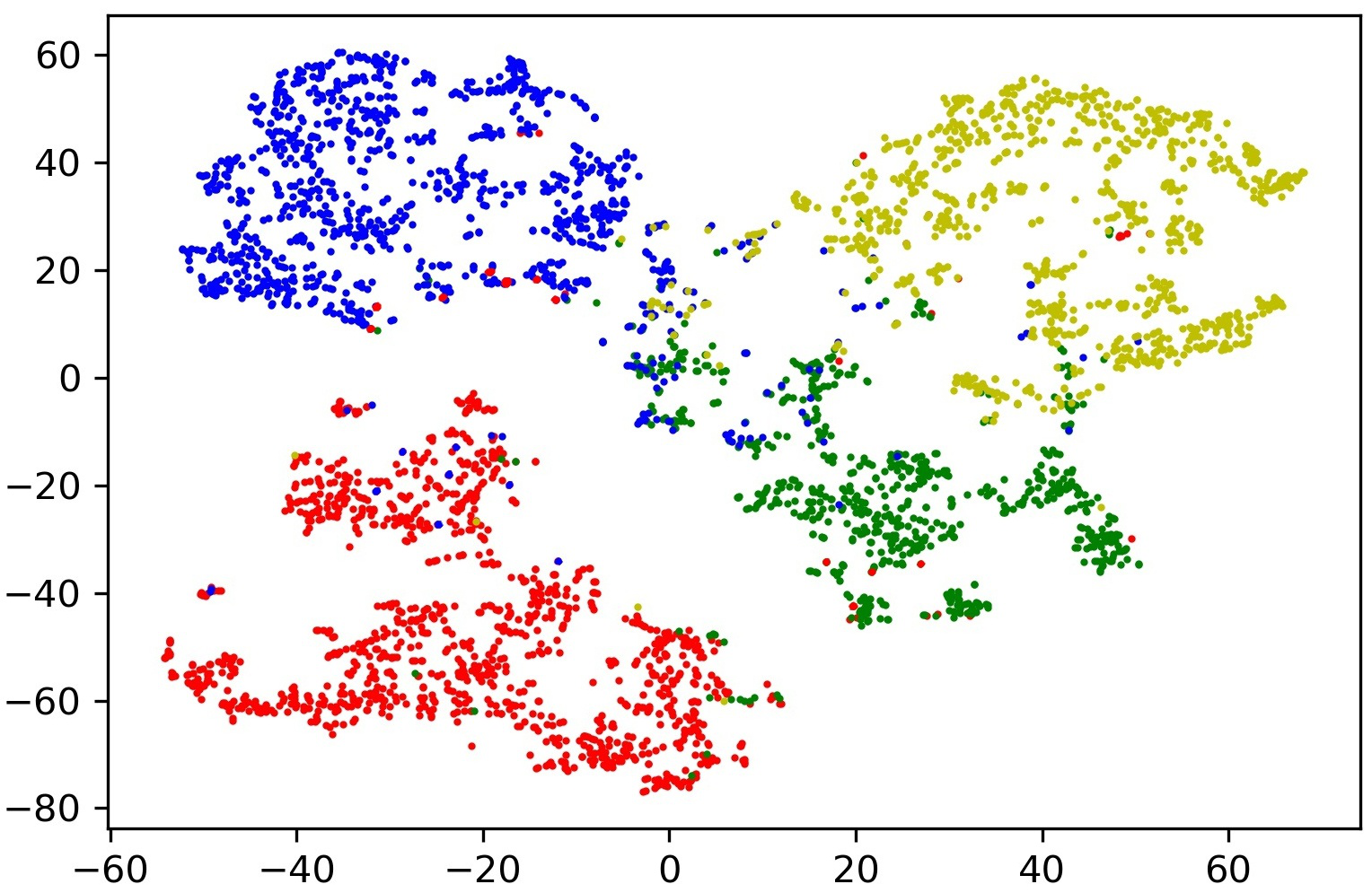}}
\end{minipage}
\begin{minipage}{.24\linewidth}
\subfloat[GAT]{\label{visual:d}\includegraphics[width=4cm]{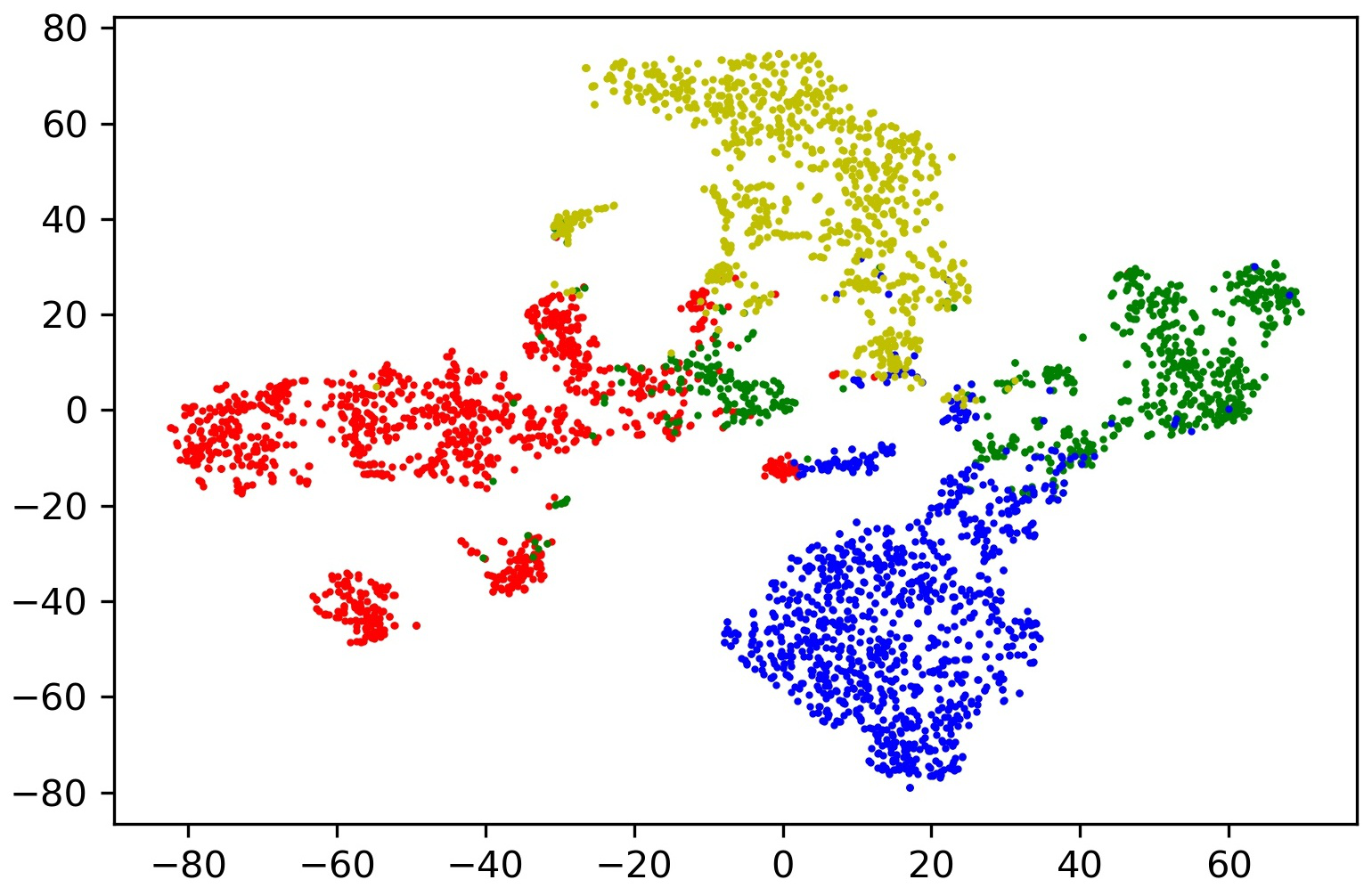}}
\end{minipage}\par\medskip
\centering
\begin{minipage}{.24\linewidth}
\subfloat[HAN]{\label{visual:e}\includegraphics[width=4cm]{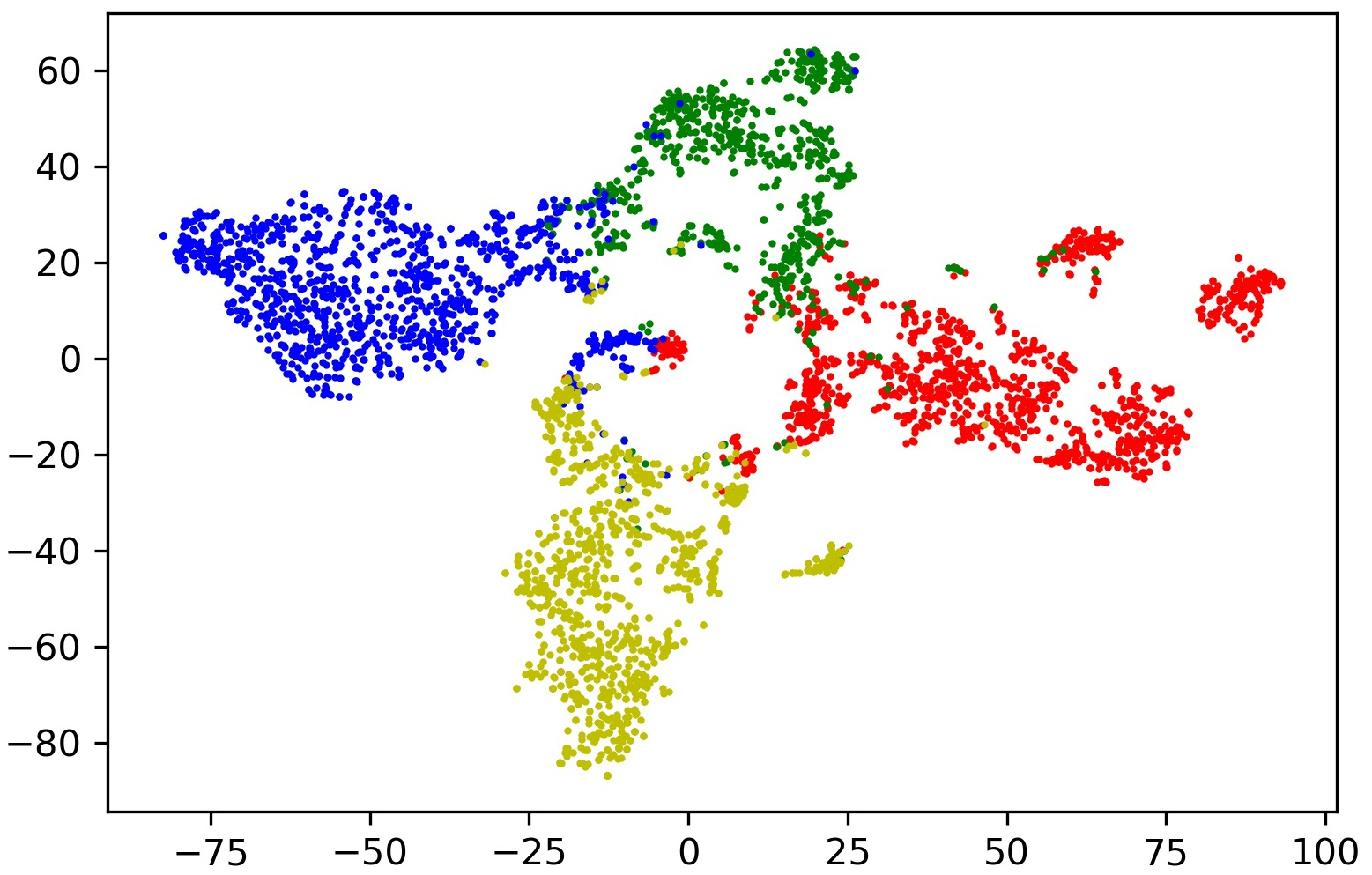}}
\end{minipage}
\begin{minipage}{.24\linewidth}
\subfloat[GTN]{\label{visual:f}\includegraphics[width=4cm]{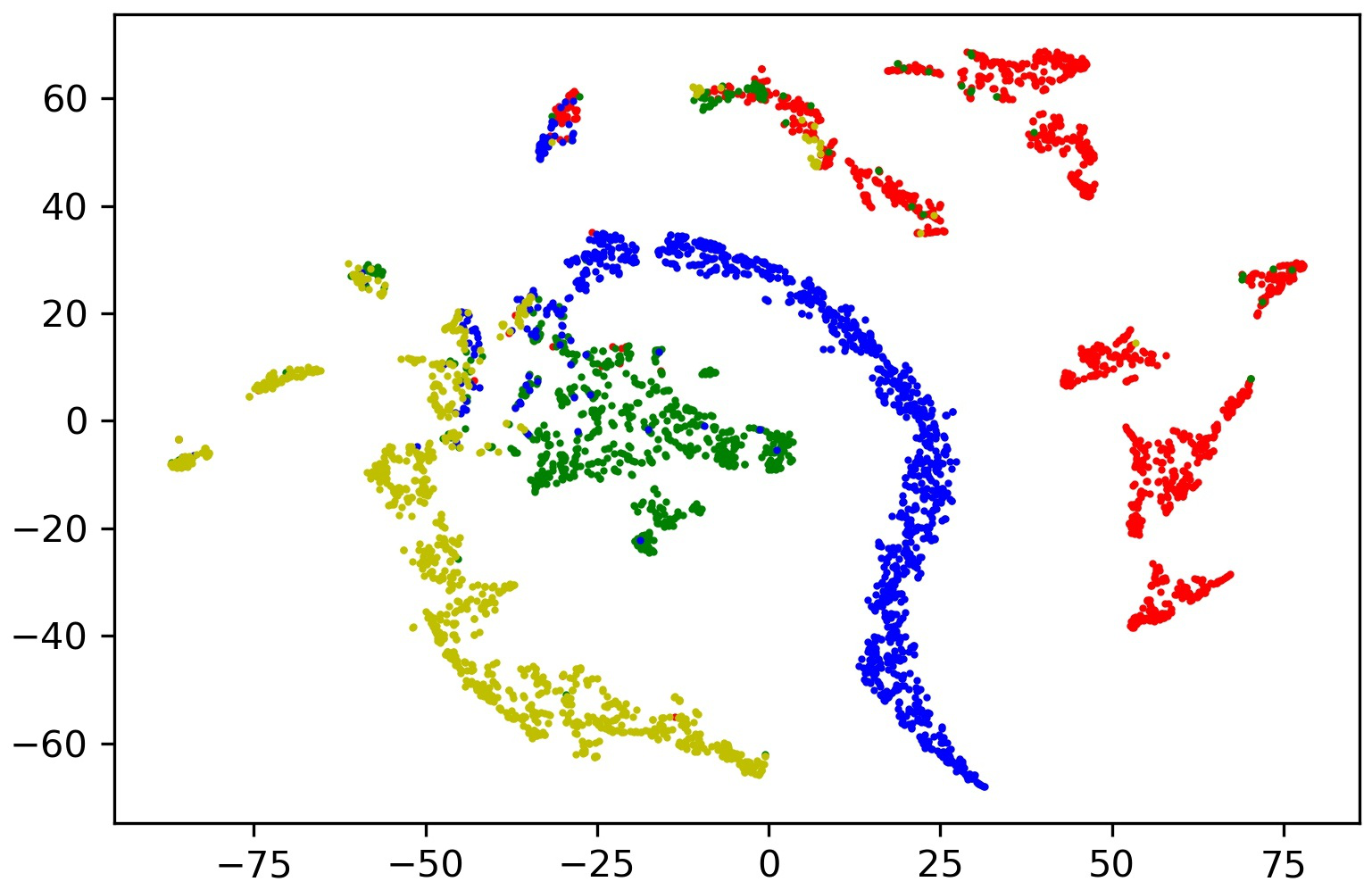}}
\end{minipage}
\begin{minipage}{.24\linewidth}
\subfloat[HetGNN]{\label{visual:k}\includegraphics[width=4cm]{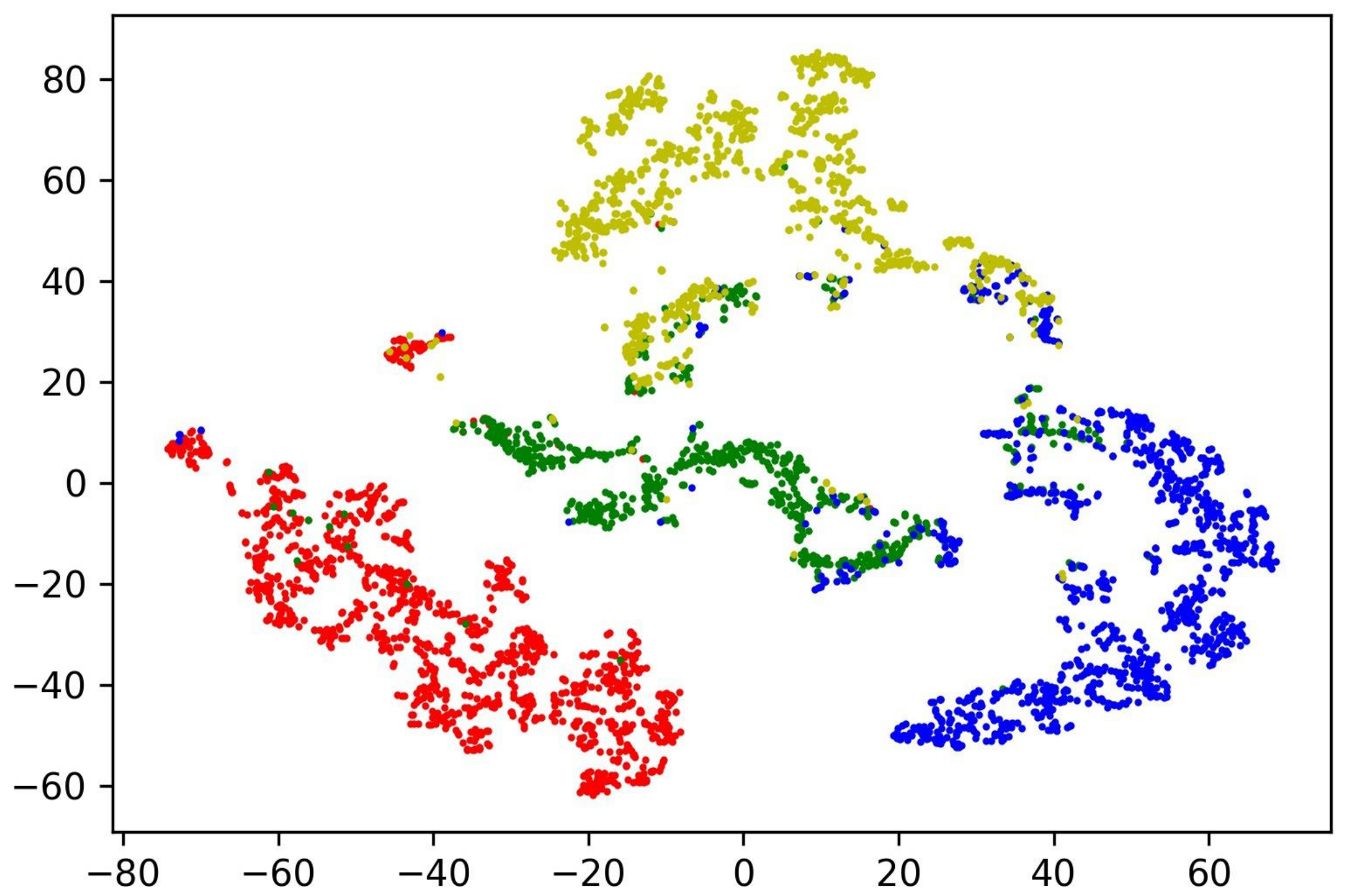}}
\end{minipage}
\begin{minipage}{.24\linewidth}
\subfloat[HGT]{\label{visual:l}\includegraphics[width=4cm]{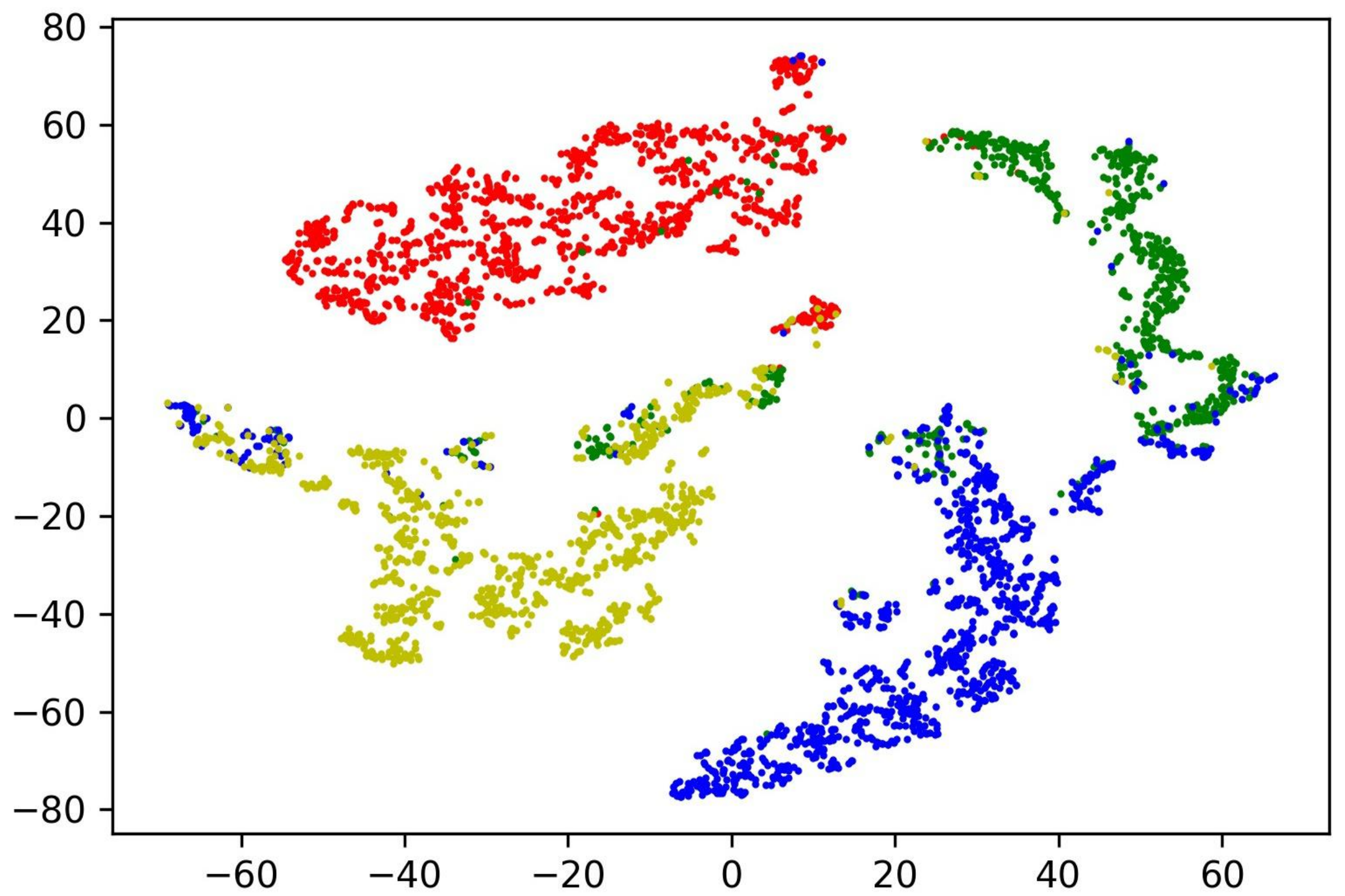}}
\end{minipage}\par\medskip
\centering
\begin{minipage}{.24\linewidth}
\subfloat[${\mathrm{HAE}_{SCL }^{2l}}$]{\label{visual:g}\includegraphics[width=4cm]{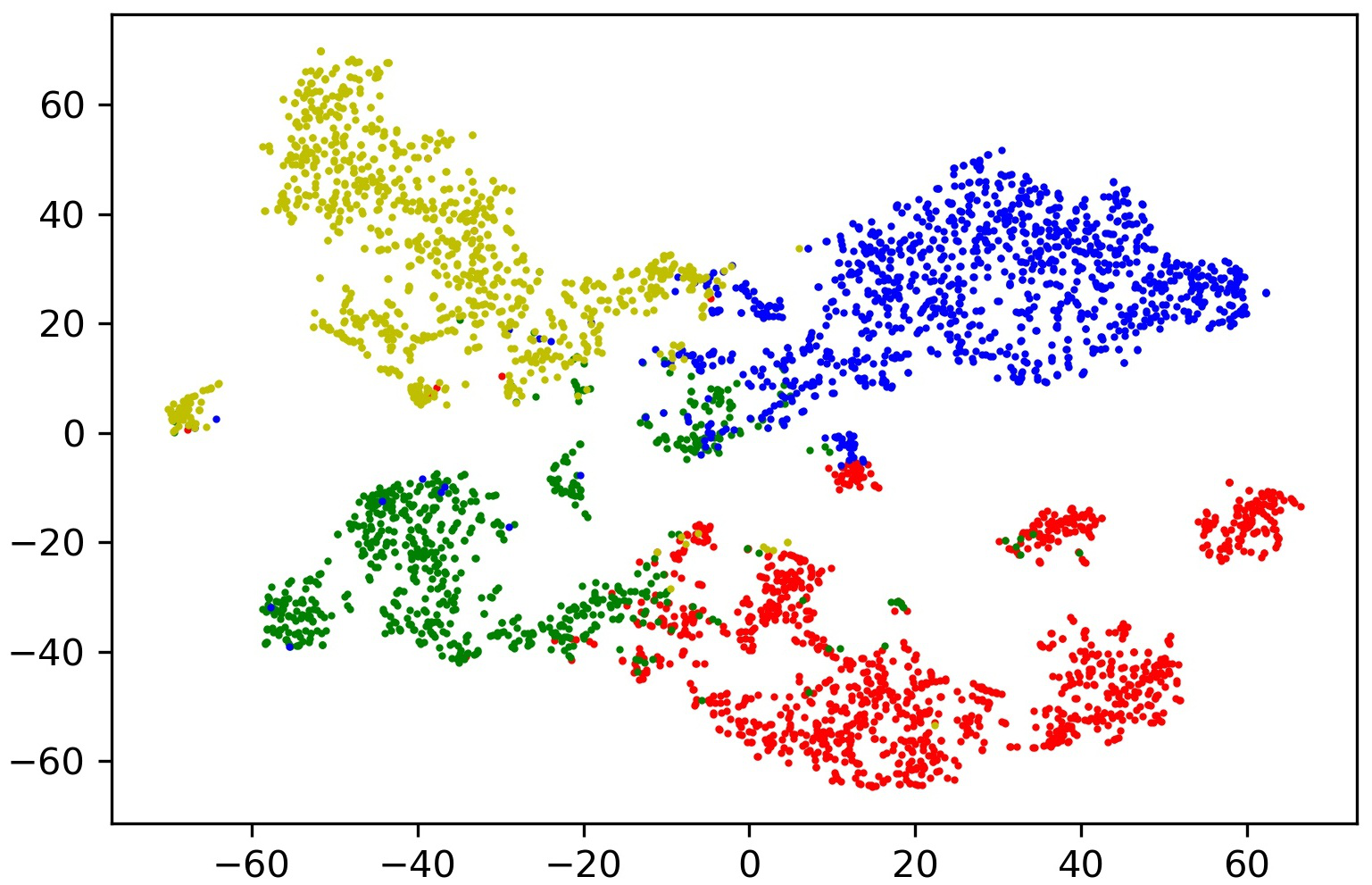}}
\end{minipage}
\begin{minipage}{.24\linewidth}
\subfloat[$\mathrm{HAE}_{GNN}^{2l}$]{\label{visual:h}\includegraphics[width=4cm]{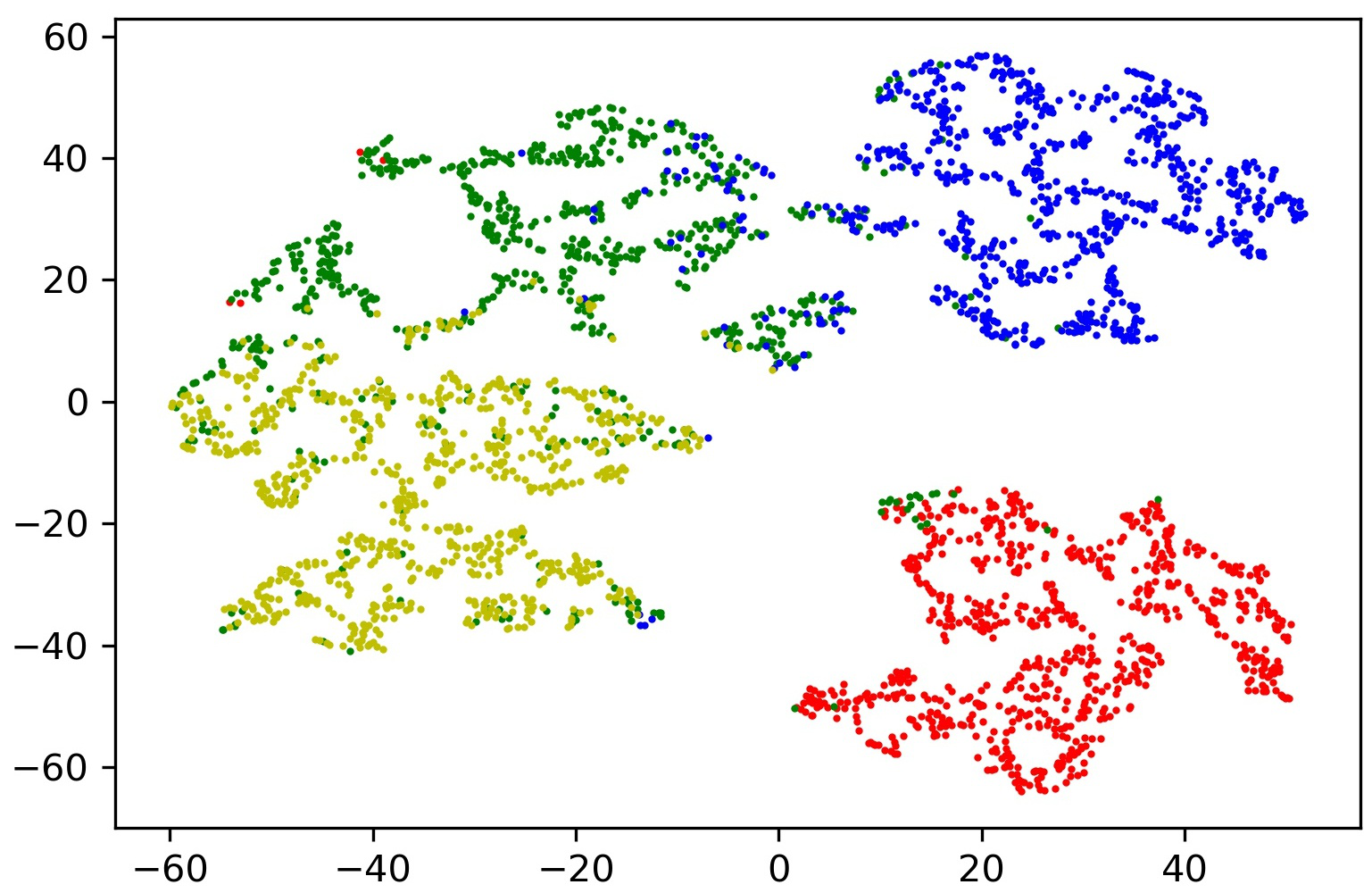}}
\end{minipage}
\begin{minipage}{.24\linewidth}
\subfloat[$\mathrm{HAE}_{CAL}^{4l}$]{\label{visual:i}\includegraphics[width=4cm]{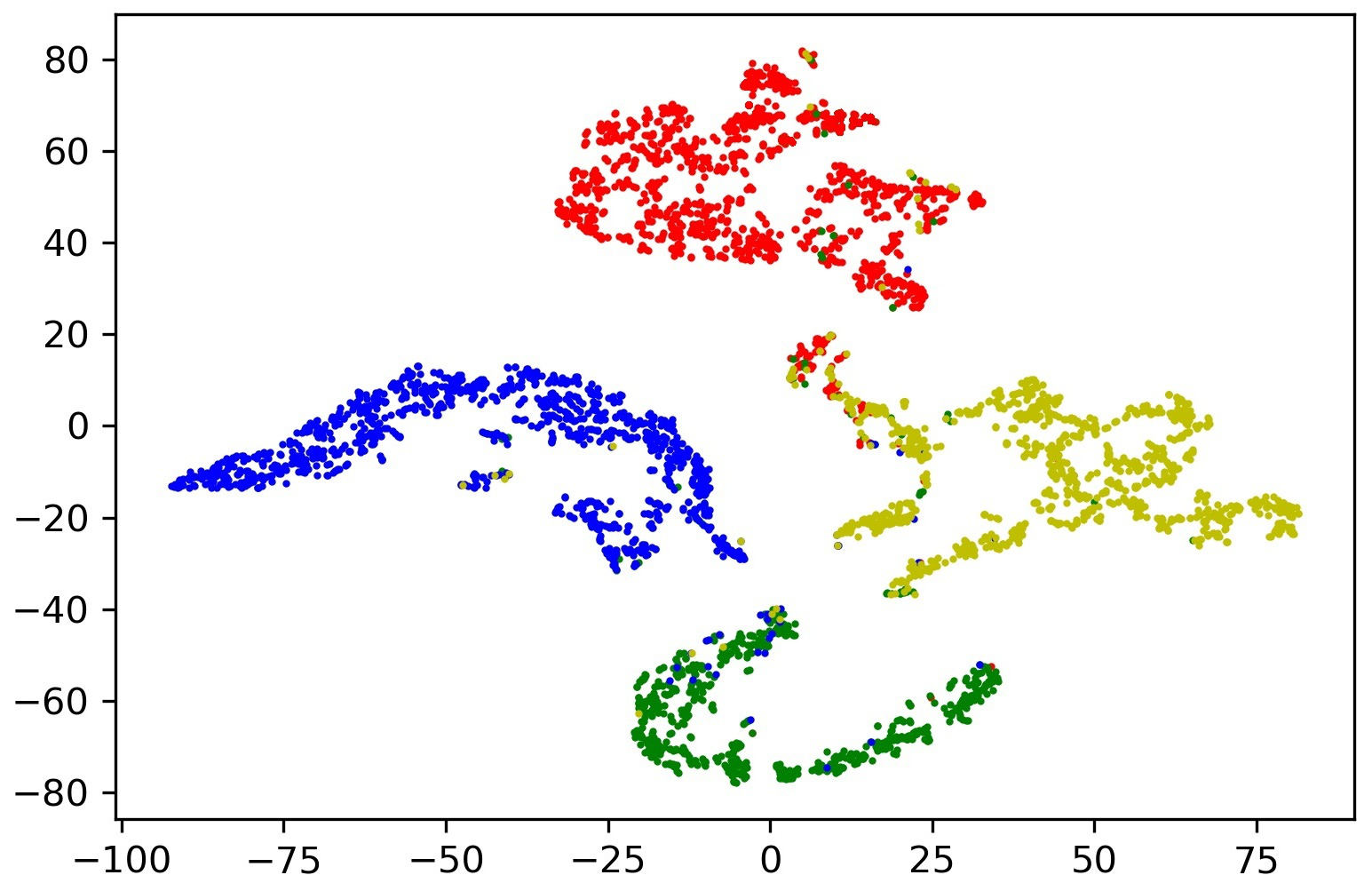}}
\end{minipage}
\begin{minipage}{.24\linewidth}
\subfloat[$\mathrm{HAE}_{GNN}^{4l}$]{\label{visual:j}\includegraphics[width=4cm]{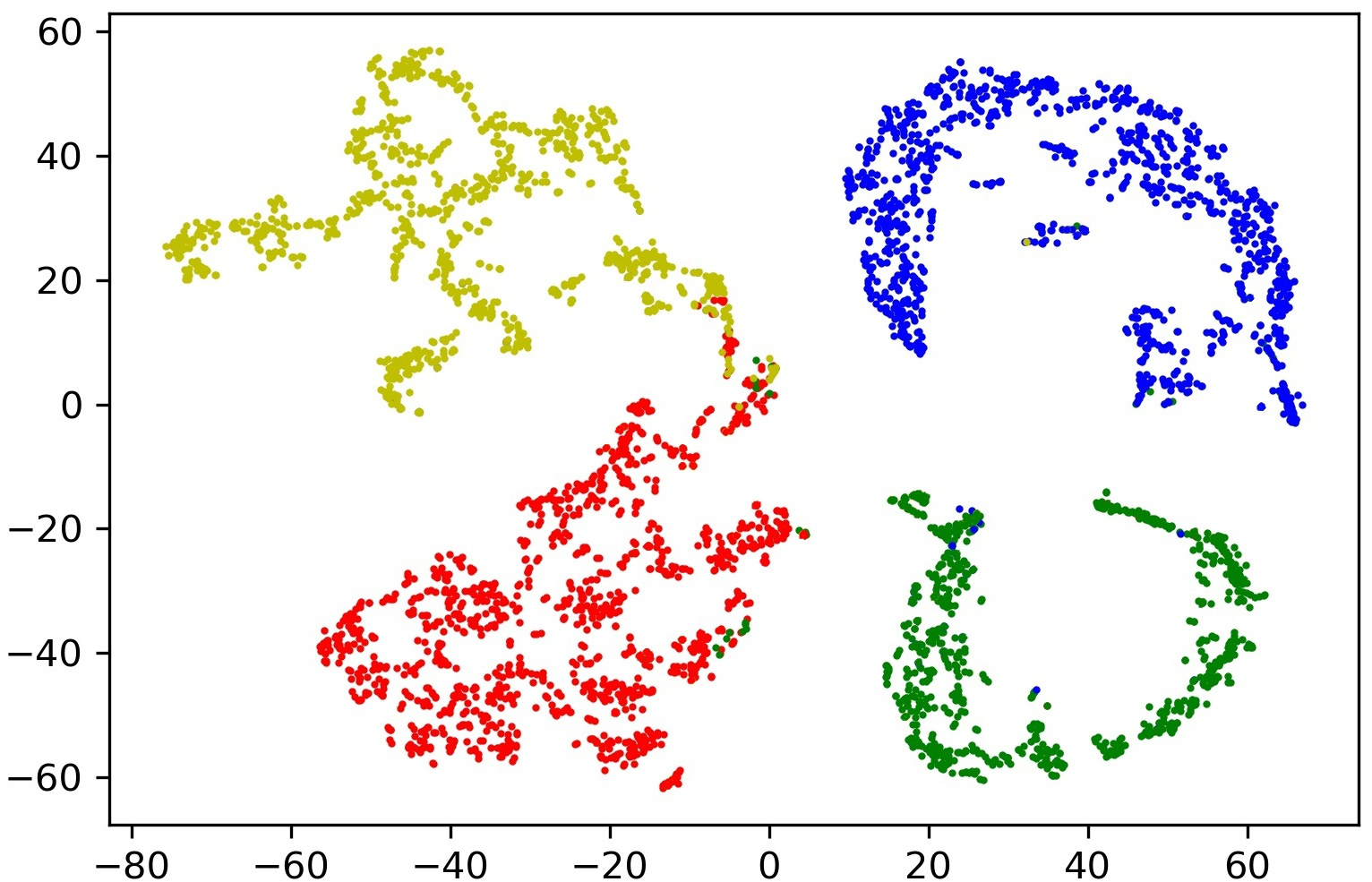}}
\end{minipage}\par\medskip\vspace{-3pt}
\caption{Visualization of the embeddings of the authors in DBLP. Each point corresponds to one author, and the color of the point indicates the author's research area.}
\label{visualization}
\end{figure*}

\subsection{Node Clustering and Visualization}\label{sec:clustering}
This section compares HAE$_{GNN}^{4l}$ to the baselines by node clustering.
Deepwalk, Metapath2vec, DANE and \FIXED{HONE} models are unsupervised.
For semi-supervised methods, including SDNE, GCN, GAT, GTN, HAN, RSHN, MAGNN, HGT, HetGNN and HAE framework based variants, the training ratio is 80\%.
\FIXED{We perform 5-fold cross-validation on each of the datasets.}
For each model, we first get the embeddings learned by it. 
We then leverage K-means for clustering and set the number of clusters to be the number of ground truth classes.
\FIXED{We report average normalized mutual information (NMI), adjusted rand index (ARI) Fowlkes-Mallows index (FMI), and standard deviations.}
We also use t-SNE \cite{maaten2008visualizing} to visualize the embeddings learned from the DBLP dataset for a direct overview.

Table~\ref{clustering_vs_base} shows the clustering results.
HAE$_{GNN}^{4l}$ consistently outperforms all baselines by large margins across the datasets, suggesting that HAE$_{GNN}^{4l}$ better captures the correlations between nodes. 
We give more comparative analysis with the representative semi-supervised GNNs.
A comparison between HAE$_{GNN}^{4l}$ and GTN shows that on the DBLP dataset, HAE$_{GNN}^{4l}$ gives $8.67\%$ higher NMI, $10.78\%$ higher ARI, and $5.70\%$ higher FMI; on the IMDB dataset, it gives $5.04\%$ higher NMI, $4.91\%$ higher ARI, $6.14\%$ higher FMI; and on the HUAWEI dataset, it gives $4.43\%$ higher NMI, $4.43\%$ higher ARI, and $4.01\%$ higher FMI. 
This is because HAE$_{GNN}^{4l}$ simultaneously captures semantic-based as well as content-based interactions between nodes, and this combined approach is more effective than the pure semantic-based approach as adopted by GTN. 
Compared to HAN, HAE$_{GNN}^{4l}$ achieves $11.13\%$ higher NMI, $9.33\%$ higher ARI, $8.70\%$ higher FMI on DBLP, achieves $7.68\%$ higher NMI, $4.77\%$ higher ARI, $10.41\%$ higher FMI on IMDB, and achieves $3.32\%$ higher NMI, $3.01\%$ higher ARI, $5.71\%$ higher FMI on HUAWEI. 
One reason is HAE$_{GNN}^{4l}$ leverages meta-paths and meta-graphs for semantics, as opposed to HAN, which solely leverages meta-paths.
Moreover, HAE$_{GNN}^{4l}$ leverages higher-order attention to fully capture nodes' interactions, as opposed to HAN, which only utilizes the first-order neighborhoods. 
In short, by leveraging the semantics contained in meta-graphs, and incorporating higher-order content-based nodes' interactions, HAE$_{GNN}^{4l}$ effectively improves the node clustering performances upon strong baselines.
\renewcommand{\arraystretch}{1.5}
\begin{table*}[t]
\caption{Node classification results (\%) of the HAE variants.} 
\label{classfication_HAE_vari}  
\centering
\fontsize{7}{8}\selectfont
\begin{tabular}{p{2.8cm}<{\centering}|p{1.8cm}<{\centering}p{1.8cm}<{\centering}|p{1.8cm}<{\centering}p{1.8cm}<{\centering}|p{1.8cm}<{\centering}p{1.8cm}<{\centering}}
\hlinewd{0.7pt}
\multirow{2}{*}{Method} & \multicolumn{2}{c}{\textbf{DBLP}}     & \multicolumn{2}{c}{\textbf{IMDB}}     & \multicolumn{2}{c}{\textbf{HUAWEI}}   \\ \cline{2-7} 
                                 & \textbf{Macro-F1} & \textbf{Micro-F1} & \textbf{Macro-F1} & \textbf{Micro-F1} & \textbf{Macro-F1} & \textbf{Micro-F1} \\ 
\hlinewd{0.7pt}
    Best Baseline & 93.15 & 93.77 & 60.01 & 60.61 & 42.69 & 43.73 \\
    
\hlinewd{0.7pt}

    HAE$_{SCL}^{2l}$ & 91.82\FIXED{$\pm$.11} & 90.86\FIXED{$\pm$.12}  & 61.44\FIXED{$\pm$.10} & 62.29\FIXED{$\pm$.09} & 41.12\FIXED{$\pm$.11} & 40.74\FIXED{$\pm$.12} \\
    HAE$_{CAL}^{4l}$ & \textit{93.07}\FIXED{$\pm$.09} & \textit{92.67}\FIXED{$\pm$.08} & \textit{61.83}\FIXED{$\pm$.08} &  \textbf{62.97}\FIXED{$\pm$.08} &  \textbf{43.28}\FIXED{$\pm$.08} & \textit{42.43}\FIXED{$\pm$.08} \\
    HAE$_{GNN}^{2l}$ & 92.30\FIXED{$\pm$.07} & 91.97\FIXED{$\pm$.06} & 61.63\FIXED{$\pm$.07} & 62.32\FIXED{$\pm$.04} & 42.37\FIXED{$\pm$.06} & 42.05\FIXED{$\pm$.05} \\
    HAE$_{GNN}^{4l}$ & \textbf{93.69}\FIXED{$\pm$.03} &  \textbf{93.82}\FIXED{$\pm$.02} &  \textbf{62.96}\FIXED{$\pm$.05} & \textit{62.61}\FIXED{$\pm$.04} & \textit{42.83}\FIXED{$\pm$.03} &  \textbf{43.97}\FIXED{$\pm$.04} \\
    \bottomrule  
\end{tabular}
\end{table*}
\renewcommand{\arraystretch}{1.5} 
\begin{table*}[t]  
  \centering  
  \fontsize{7}{8}\selectfont  
  \caption{Node clustering results (\%) of the HAE variants.}  
  \label{clustering_HAE_vari}  

    \begin{tabular}{p{2.8cm}<{\centering}|p{1.2cm}<{\centering}p{1.2cm}<{\centering}p{1.2cm}<{\centering}|p{1.2cm}<{\centering}p{1.2cm}<{\centering}p{1.2cm}<{\centering}|p{1.2cm}<{\centering}p{1.2cm}<{\centering}p{1.2cm}<{\centering}}
    \hlinewd{0.7pt}
    \multirow{2}{*}{Method} &
    
    \multicolumn{3}{c}{\textbf{DBLP}} & \multicolumn{3}{c}{\textbf{IMDB}} &    \multicolumn{3}{c}{\textbf{HUAWEI}} \\ \cline{2-10}
    
    &\textbf{NMI}&\textbf{ARI}&\textbf{FMI}&\textbf{NMI}&\textbf{ARI}&\textbf{FMI}&\textbf{NMI}&\textbf{ARI}&\textbf{FMI}\\
    
    \hlinewd{0.7pt}
    Best Baseline & 73.00 & 77.48 & 84.85 & 30.86 & 29.59 & 42.03 & 29.25 & 27.24 & 37.37 \cr  
    
    \hlinewd{0.7pt}
    
    HAE$_{SCL}^{2l}$ & 74.28\FIXED{$\pm$.11} & 77.62\FIXED{$\pm$.10} & 83.26\FIXED{$\pm$.11} & 28.76\FIXED{$\pm$.09} & 28.13\FIXED{$\pm$.08} & 41.29\FIXED{$\pm$.07} & 28.26\FIXED{$\pm$.17} & 25.63\FIXED{$\pm$.12} & 35.95\FIXED{$\pm$.14} \cr
    HAE$_{CAL}^{4l}$ & \textit{76.96}\FIXED{$\pm$.08} & \textit{79.33}\FIXED{$\pm$.09} & \textit{85.56}\FIXED{$\pm$.09} & \textit{31.33}\FIXED{$\pm$.07} & \textbf{32.12}\FIXED{$\pm$.09} & \textbf{45.58}\FIXED{$\pm$.09} & \textbf{30.33}\FIXED{$\pm$.06} & \textit{26.82}\FIXED{$\pm$.09} & \textit{38.24}\FIXED{$\pm$.08} \cr
    HAE$_{GNN}^{2l}$ & 75.42\FIXED{$\pm$.05} & 78.64\FIXED{$\pm$.05} & 84.03\FIXED{$\pm$.05} & 29.36\FIXED{$\pm$.04} & 28.54\FIXED{$\pm$.03} & 42.51\FIXED{$\pm$.05} & 28.32\FIXED{$\pm$.05} & 25.13\FIXED{$\pm$.06} & 36.28\FIXED{$\pm$.05} \cr
    HAE$_{GNN}^{4l}$ & \textbf{77.64}\FIXED{$\pm$.02} & \textbf{79.79}\FIXED{$\pm$.03} & \textbf{85.62}\FIXED{$\pm$.03} & \textbf{32.63}\FIXED{$\pm$.02} & \textit{31.28}\FIXED{$\pm$.03} & \textit{44.82}\FIXED{$\pm$.02} & \textit{30.25}\FIXED{$\pm$.03} & \textbf{28.17}\FIXED{$\pm$.03} & \textbf{38.92}\FIXED{$\pm$.02}\cr
    \bottomrule  
    \end{tabular}  
\end{table*} 
Compared to RSHN, HAE$_{GNN}^{4l}$ achieves $11.55\%$ improvements in terms of NMI, $9.91\%$ in terms of ARI, $9.52\%$ in terms of FMI on DBLP, $8.12\%$ improvements in terms of NMI, $4.86\%$ in terms of ARI, $10.67\%$ in terms of FMI on IMDB, and $3.18\%$ improvements in terms of NMI, $2.70\%$ in terms of ARI, $4.19\%$ in terms of FMI on HUAWEI.
Compared to MAGNN, HGT and HetGNN, HAE$_{GNN}^{4l}$ also achieves at least $4.64\%$ improvements in terms of NMI, $2.35\%$ in terms of ARI, $1.07\%$ in terms of FMI on DBLP, $1.77\%$ improvements in terms of NMI, $1,69\%$ in terms of ARI, $2.79\%$ in terms of FMI on IMDB, and $1.00\%$ improvements in terms of NMI, $0.93\%$ in terms of ARI, $1.55\%$ in terms of FMI on HUAWEI. 
This suggests that the proposed higher-order attribute-enhancing approach is more effective in node clustering tasks.

Figs. \ref{visualization}(a)-(h) present the visualization results of the baseline models, and (l) that of HAE$_{GNN}^{4l}$ on the DBLP dataset.
The visualization result of HAE$_{GNN}^{4l}$ is significantly better than the baselines. 
Specifically, the number of the clusters is consistent with the number of the ground truth classes, the clusters are compact, and the separations between the clusters are clear. 
This qualitatively verifies that HAE$_{GNN}^{4l}$ better captures the relations between nodes compared to the baselines.
As expected, metapath2vec gives a better visualization result than the homogeneous Deepwalk model, because it introduces heterogeneity.
However, the boundaries of its clusters are fairly blurry. 
The GNN based models perform better than the unsupervised random-walk based models, in the sense that the clusters are much clearly separated. 
Also, compared to GCN and GAT, HAN, GTN, HetGNN and HGT incorporate richer semantics, and as a result, their results show fewer overlaps between clusters.

\subsection{Comparison of the HAE Variants}\label{sec:HAE_comp}
This section compares different variants under the proposed HAE framework. 
We experiment with three architectures, i.e., HAE$_{GNN}$, HAE$_{SCL}$ and HAE$_{CAL}$. 
Specifically, we study HAE$_{GNN}^{2l}$ and HAE$_{GNN}^{4l}$ for HAE$_{GNN}$, HAE$_{SCL}^{2l}$ for HAE$_{SCL}$, and HAE$_{CAL}^{4l}$ for HAE$_{CAL}$. 
We conduct node classification, node clustering and visualization for a comprehensive comparison.
The default training ratio is 80\%, and other experimental settings are conducted in the same manner as in Sections \ref{sec:classification} and \ref{sec:clustering}.
\FIXED{We perform 5-fold cross-validation on each of the datasets.}

We present the node classification results and \FIXED{standard deviations} of the HAE models in Table~\ref{classfication_HAE_vari}. 
We copy the highest results given by the baselines from Table~\ref{classfication_vs_base} for a direct comparison. 
We see that the heterogeneous graph neural network models, including HGT and HetGNN, have achieved better node classification performances than HAE$_{SCL}^{2l}$, HAE$_{GNN}^{2l}$ and HAE$_{CAL}^{4l}$, but they are all lower than the performances of HAE$_{GNN}^{4l}$.
The experimental results of consistency improvement show that the semantic-based and content-based nodes' interactions and higher-order attention architecture are all directly conducive to give better performances.
HAE$_{GNN}^{4l}$ outperforms HAE$_{CAL}^{4l}$ in four out of six cases, and outperforms HAE$_{GNN}^{2l}$ in all the cases, which suggests that the rich semantics and the higher-order architecture both contribute to better classification performances, and should be used in combination. 
\FIXED{Even as the number of layers increases, the performance of the model becomes more stable.}

Table~\ref{clustering_HAE_vari} presents the node clustering results and \FIXED{standard deviations} of the HAE models, along with the highest baselines' results copied from Table \ref{clustering_vs_base}. 
Most variants of HAE are better than the best baseline results, and the HAE$_{GNN}^{4l}$ model has achieved consistent, significant \FIXED{and stable} improvements compared to the best baseline results.
HAE$_{GNN}^{4l}$ and HAE$_{CAL}^{4l}$ perform better than the other two HAE models, and HAE$_{GNN}^{4l}$ outperforms HAE$_{CAL}^{4l}$ in six out of nine cases. 
This testifies the effectiveness of combining semantics with higher-order architectures. 
Figs.~\ref{visualization}(i)-(l) present the visualization results of the HAE models. 
HAE$_{GNN}^{4l}$ and HAE$_{CAL}^{4l}$ show better visualization results than the other two, in the sense that the clusters are more compact and the separations are clearer. 
This suggests that a higher order attribute-enhancing framework can help improve the visualization performances.

To conclude, experimental results imply the effectiveness of our semantic incorporation and higher-order attribute-enhancing approaches. 
We provide a more detailed analysis of their effects in Sections \ref{sec:sema_eval} and \ref{sec:higher_eval}. 
HAE$_{GNN}^{4l}$, which leverages these approaches in combination, is generally a better performing one among the HAE models.

\renewcommand{\arraystretch}{1.5} 
\begin{table*}[t]  
  \centering  
  \fontsize{7}{8}\selectfont  
  \caption{Comparison of node classification results (\%) with and without using meta-graphs. Macro/Micro-F1 is the result without using meta-graphs, and $\Delta$ means the difference between using meta-graphs and not using them.}  
  \label{classification_wo_metagraphs}  
    \begin{tabular}{p{2.5cm}<{\centering}|p{0.9cm}<{\centering}p{0.9cm}<{\centering}p{0.9cm}<{\centering}p{0.9cm}<{\centering}|p{0.9cm}<{\centering}p{0.9cm}<{\centering}p{0.9cm}<{\centering}p{0.9cm}<{\centering}|p{0.9cm}<{\centering}p{0.9cm}<{\centering}p{0.9cm}<{\centering}p{0.9cm}<{\centering}}  
    \hlinewd{0.7pt}
    \multirow{2}{*}{Method} &
    
    \multicolumn{4}{c}{\textbf{DBLP}} & \multicolumn{4}{c}{\textbf{IMDB}} &    \multicolumn{4}{c}{\textbf{HUAWEI}} \\ \cline{2-13}
    
    &\textbf{Macro-F1}&\textbf{$\Delta$Macro-F1}&\textbf{Micro-F1}&\textbf{$\Delta$Micro-F1}&\textbf{Macro-F1}&\textbf{$\Delta$Macro-F1}&\textbf{Micro-F1}&\textbf{$\Delta$Micro-F1}&\textbf{Macro-F1}&\textbf{$\Delta$Macro-F1}&\textbf{Micro-F1}&\textbf{$\Delta$Micro-F1}\cr
    
    \hlinewd{0.7pt}
    Best Baseline  & 93.15 & - & 93.77  & - & 60.01 & - & 60.61 & - & 42.69 & - & 43.73 & - \cr  
    
    \hlinewd{0.7pt}
    HAE$_{SCL}^{2l}$ & 90.76\FIXED{$\pm$.08} & -1.06 & 90.33\FIXED{$\pm$.08} & -0.53 & 60.17\FIXED{$\pm$.07} & -1.27 & 60.92\FIXED{$\pm$.07} & -1.37 & 39.12\FIXED{$\pm$.06} & -2 & 39.38\FIXED{$\pm$.06} & -1.36\cr
    HAE$_{CAL}^{4l}$ & \textbf{92.76}\FIXED{$\pm$.07} & -0.31 & 92.17\FIXED{$\pm$.06} & -0.5 & 61.04\FIXED{$\pm$.07} & -0.79 & 61.24\FIXED{$\pm$.06} & -1.73 & \textbf{42.46}\FIXED{$\pm$.06} & -0.82 & 42.13\FIXED{$\pm$.05} & -0.3\cr
    HAE$_{GNN}^{2l}$ & 91.25\FIXED{$\pm$.06} & -1.05 & 91.37\FIXED{$\pm$.05} & -0.6 & 60.42\FIXED{$\pm$.05} & -1.21 & 61.27\FIXED{$\pm$.05} & -1.05 & 40.18\FIXED{$\pm$.04} & -2.19 & 40.37\FIXED{$\pm$.04} & -1.68\cr
    HAE$_{GNN}^{4l}$ & 92.55\FIXED{$\pm$.04} & -1.14 & \textbf{92.95}\FIXED{$\pm$.03} & -0.87 & \textbf{62.38}\FIXED{$\pm$.04} & -0.58 & \textbf{62.09}\FIXED{$\pm$.03} & -0.52 & 41.83\FIXED{$\pm$.03} & -1 & \textbf{42.54}\FIXED{$\pm$.02} & -1.43\cr
    \bottomrule  
    \end{tabular}  
\end{table*}

\begin{figure*}[h]
    \centering
    {
    \begin{minipage}[c]{0.3\textwidth}
    \centering
        \includegraphics[width =5.2cm]{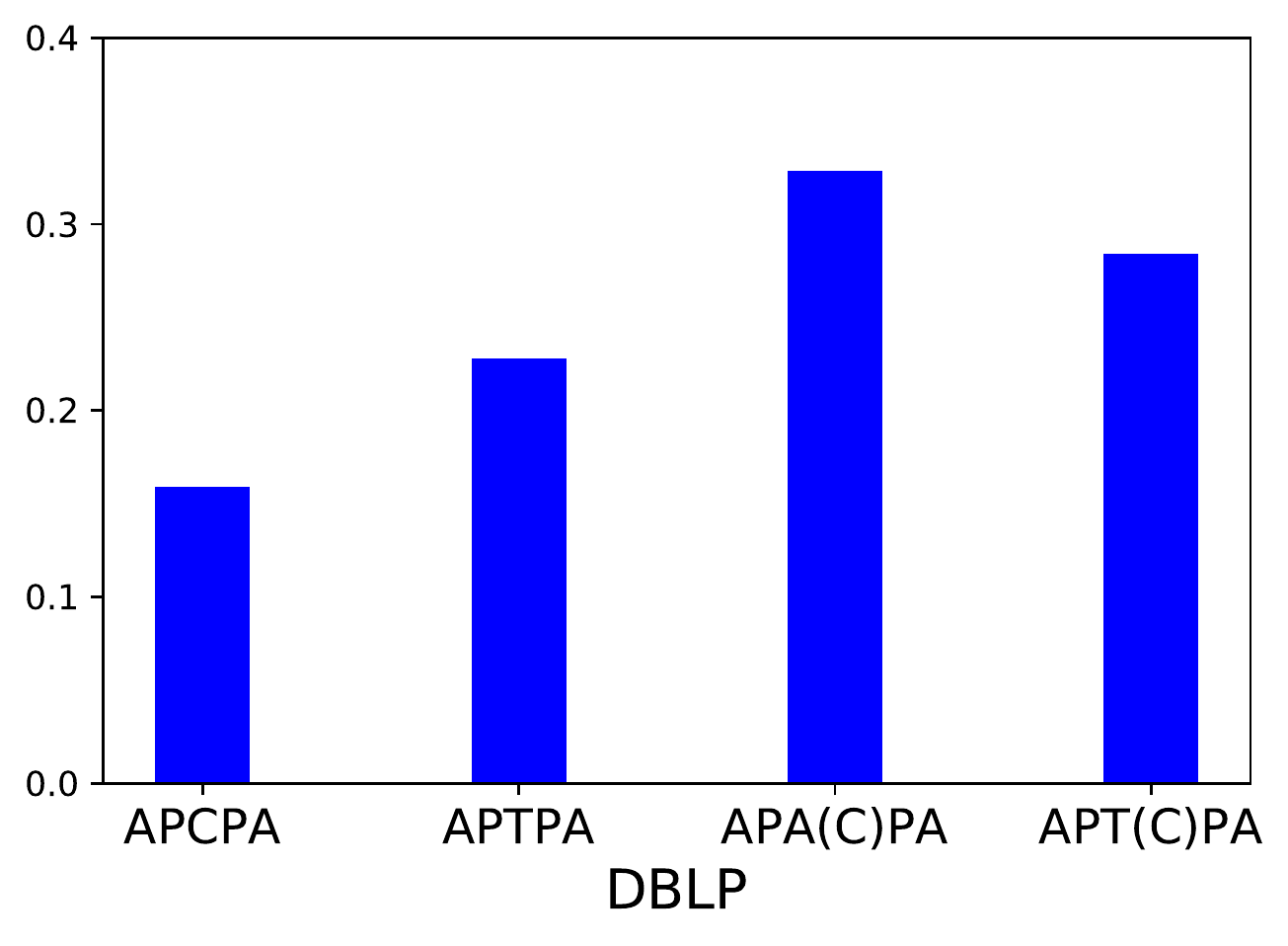}
    \end{minipage}
    }\vspace{-5pt}
    \hfill
    {
    \begin{minipage}[c]{0.3\textwidth}
    \centering
        \includegraphics[width =5.2cm]{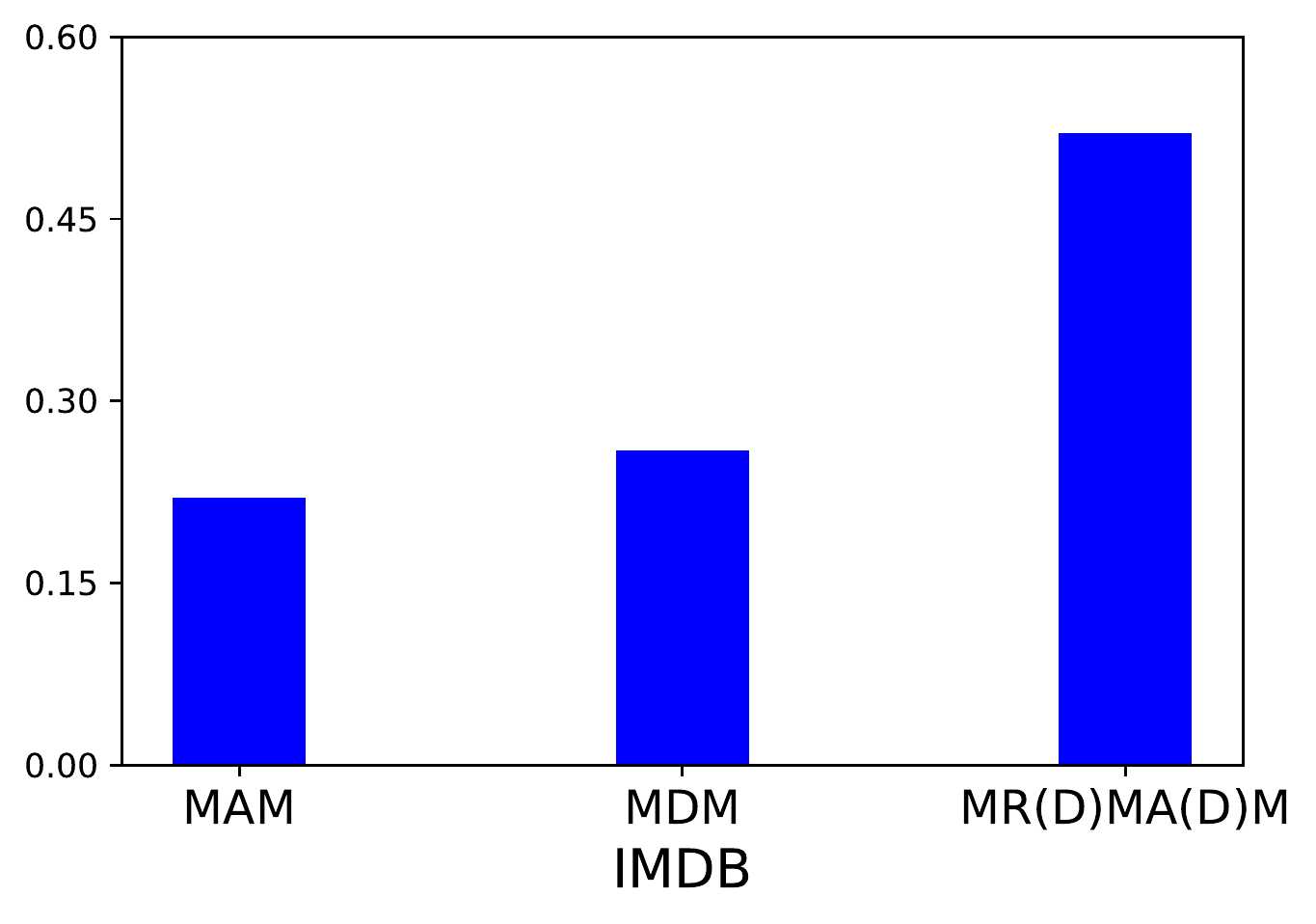}
    \end{minipage}
    }\vspace{-5pt}
    \hfill
    {
    \begin{minipage}[c]{0.3\textwidth}
    \centering
        \includegraphics[width =5.2cm]{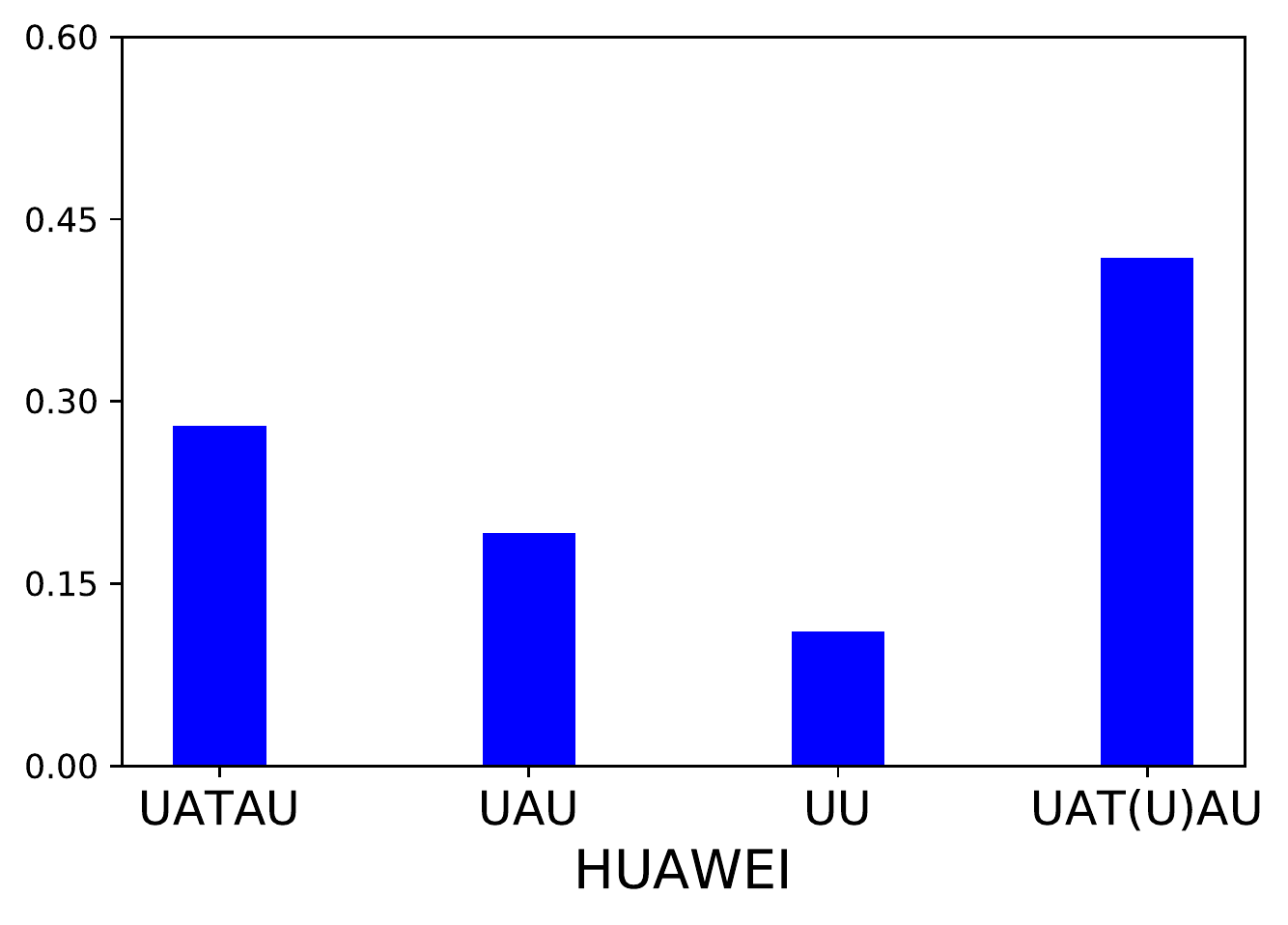}
    \end{minipage}
    }\vspace{-5pt}
    \caption{The weights of the semantic structures learned by HAE$_{GNN}^{4l}$.}
    \label{weights}
\end{figure*}

\subsection{Evaluation of Semantic Incorporation}\label{sec:sema_eval}
This section evaluates the effectiveness of our semantic incorporation.
Recall that our approach is novel in two aspects.
First, in addition to meta-paths, we learn semantics from meta-graphs, which are omitted by previous heterogeneous GNNs.
Second, we associate a trainable weight with each semantic structure to emphasize more on the important semantic structures.
To evaluate the former, we conduct node classification experiments without using the meta-graphs for a comparison.
To evaluate the latter, we present and interpret the learned weights of the meta-paths and meta-graphs used in this study.

We reconstruct the HAE models, including HAE$_{GNN}^{4l}$, HAE$_{GNN}^{2l}$, HAE$_{SCL}^{2l}$ and HAE$_{CAL}^{4l}$ solely using the meta-paths. 
The results are presented in Table \ref{classification_wo_metagraphs}. 
\FIXED{Besides Marco-F1, Micro-F1 and standard deviations}, we also present the drops in these two metrics compared to the original node classification results shown in Table \ref{classfication_vs_base}.
We copy the highest results given by the baselines from Table \ref{classfication_vs_base} for a direct comparison.
We can tell that incorporating meta-graphs can help improve node classification performances, as all our models give higher results when simultaneously leveraging meta-paths and meta-graphs, and removing the meta-graphs could result in a drop of $0.3\%\sim2.2\%$ in Macro-F1 and $0.3\%\sim1.7\%$ in Micro-F1. 
This indicates that incorporating meta-graphs in the learning process is quite beneficial.

\begin{figure*}
    \centering
    \includegraphics[width=18cm]{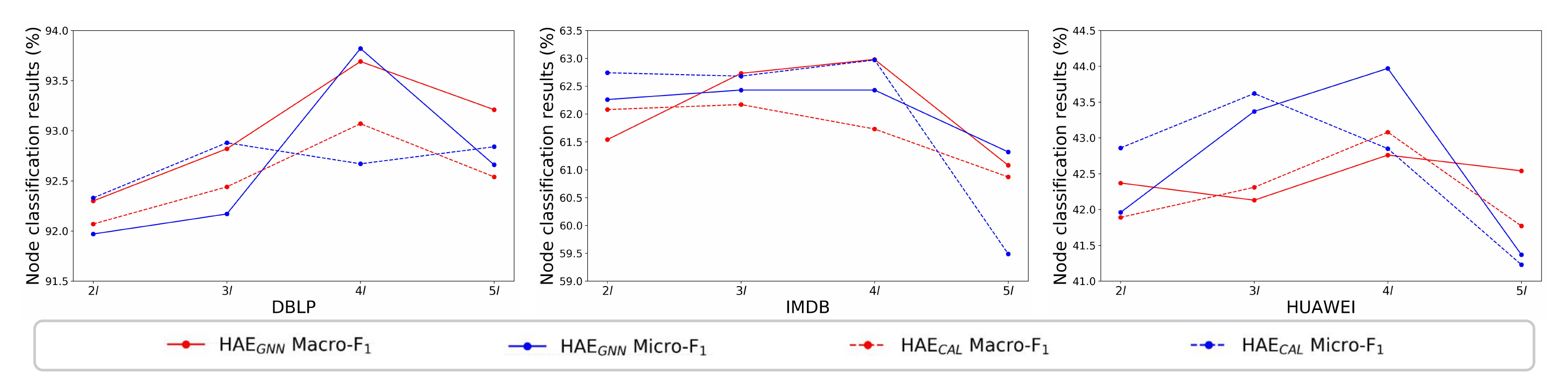}\vspace{-10pt}
    \caption{Node classification results with different orders. Here we experiment on HAE$_{GNN}$ and HAE$_{CAL}$ of order 2 through 5. For each model, we conduct node classification on all the three datasets, and present Macro-F1 as well as Micro-F1 scores.}
    \label{casastudy}
\end{figure*}

\begin{figure*}
    \centering
    \includegraphics[width=18cm]{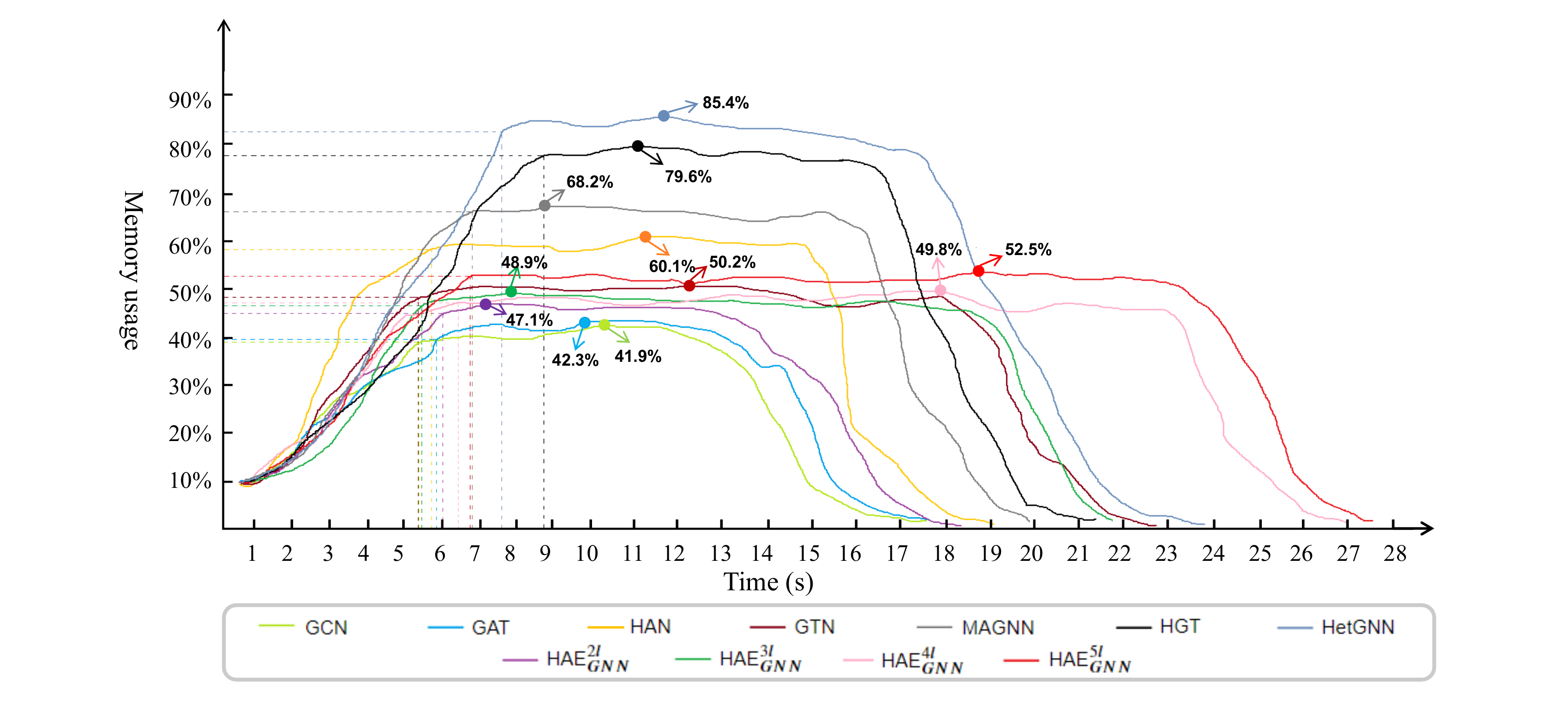}\vspace{-10pt}
    \caption{Memory consumptions over time. We run each model for 100 epochs on the DBLP dataset. The percentage of memory usage is measured at 8 GB. We mark the maximum memory consumption of each model.}
    \label{efficiency}
\end{figure*}

We report the weights of all semantic structures used in this study that are learned by the proposed HAE$_{GNN}^{4l}$ model in Fig.~\ref{weights}. 
In general, for all the three datasets, the meta-graphs have higher weights than the meta-paths. 
This suggests that the model relies more on the meta-graphs rather than the meta-paths. 
This is reasonable, as the meta-graphs convey more complex semantics, and are better indicators of the nodes' relatedness. 
For example, in DBLP, meta-graph $APT(C)PA$ is more rigid than meta-path $APCPA$, and therefore, two authors connected by the former are more likely to be in the same research field as opposed to those that are connected by the latter. 
We also compare the different meta-paths. 
In DBLP, meta-path $APTPA$ has a higher weight than $APCPA$, because working on a common subject can better indicate the similarity between two authors' research fields as compared to submitting to a common conference. 
In IMDB, $MDM$ has a slightly higher weight than $MAM$, as directors are more likely to devote themselves to specific movie genre types than the actors do. 
In HUAWEI, $UATAU$ has a higher weight than the other two meta-paths. This is reasonable. 
For example, two users that both use English learning apps are likely to be students of similar ages. 
In contrast, $UU$ has a lower weight, because age is not the main factor that affects people when choosing who to follow on social media. 
In essence, through the use of the weights vector, the proposed model is capable of relying more on the important semantic structures.

To conclude, the proposed HAE$_{GNN}$ model leverages meta-graphs, which, as shown in the experiments, help enhance the performances.
HAE$_{GNN}$ also detects the important semantic structures and learns interpretable weights for all semantic structures.

\subsection{Evaluation of Higher-Order Attribute-Enhancing}\label{sec:higher_eval}
In this section, we study the effects of the higher-orderliness in the proposed HAE framework.

We first observe how the order of our models affects the node classification performances. 
We experiment with two models under the HAE framework, i.e., HAE$_{GNN}$ and HAE$_{CAL}$. 
We let the orders of them vary from 2 to 5, and present their node classification performances in Fig.~\ref{casastudy}.
On all the three datasets, the Macro-F1 and Micro-F1 scores of HAE$_{GNN}$ generally increase as we increase the order up to 4.
This is because as the order becomes higher, HAE$_{GNN}$ gradually incorporates more CALs, and pays attention to more distant neighbors of each node. 
The performances of HAE$_{GNN}$ reach the maximum when the order reaches 4, and start to drop after that because of overfitting. 
For HAE$_{CAL}$, higher order also helps, but its effects are not as significant. 
On DBLP, HAE$_{CAL}^{3l}$ shows higher Macro-F1 and Micro-F1 scores than HAE$_{CAL}^{2l}$, and the HAE$_{CAL}^{4l}$ performs slightly better than HAE$_{CAL}^{3l}$ if considering the averages of their Macro-F1 and Micro-F1 scores. 
On IMDB and HUAWEI, varying the order does not affect the performances by much, as HAE$_{CAL}^{2l}$, HAE$_{CAL}^{3l}$ and HAE$_{CAL}^{4l}$ show around the same average F1 scores.

The performances of HAE$_{CAL}$ also start to drop, as the effect of overfitting, when the order becomes too high ($>$4). 
Also, a comparison between the datasets suggests that the higher-order models are more effective for the DBLP dataset, as HAE$_{GNN}^{4l}$ and HAE$_{CAL}^{4l}$ improve upon HAE$_{GNN}^{2l}$ and HAE$_{CAL}^{2l}$, respectively. 
On HUAWEI and IMDB, however, the effects of increasing the order are less significant. 
In short, higher order can help improve the node classification performances, and it is especially effective for the proposed HAE$_{GNN}$ model. 
The effectiveness of the higher-orderliness is also affected by the dataset.

We also observe the memory efficiency of HAE$_{GNN}$ when its order varies.
Fig.~\ref{efficiency} shows the changes in memory usage of HAE$_{GNN}$ models as well as the GNN based baseline models running on the DBLP dataset for 100 epochs. 
Here, all processes of computing the commuting matrix are performed in parallel on GPUs, and are included in the Fig.~\ref{efficiency}.
It is worth noting that HAE$_{GNN}^{2l}$, which outperforms the previous state-of-the-art baselines such as HAN, GTN in node classification and node clustering, is also more efficient in terms of memory and/or time consumptions. 
Fig.~\ref{efficiency} shows that HAE$_{GNN}^{2l}$ requires around $19.6\%$ and $7.4\%$ less memory as compared to HAN and GTN, respectively. 
Furthermore, HAE$_{GNN}^{2l}$ takes less time than GTN (20 versus 23 seconds) to run for 100 epochs. 
Compared to GCN, models that involve multi-head attention such as GAT, HAN, HGT, MAGNN and HAE$_{GNN}$ models, or multiple channels such as GTN, require more memory. 
Among them, HGT consumes more memory than the others, because it not only applies multi-head attention units but also uses residual neural networks, message passing, aggregation functions, etc. 
Among all methods, HetGNN consumes the most memory, because it simultaneously utilizes multi-head attention units, bi-LSTM units and other complex neural networks.
HAE$_{GNN}$ models require more memory as the order increases; however, the increments in memory consumption are not significant.
Also, higher-order HAE$_{GNN}$ models take more time to run because of their deeper structures.
As the number of layers increases, the time consumption of the HAE$_{GNN}$ models also increases, which is an acceptable training process.

To conclude, the higher-order architecture in HAE$_{GNN}$ helps improve its performance. 
HAE$_{GNN}$ is able to achieve superior performances with less memory and/or time consumptions compared to state-of-the-art baselines.

\section{Conclusion And Future Work}\label{sec:conclusion}
In this paper, we propose a novel HAE$_{GNN}$ model for heterogeneous network representation learning. 
HAE$_{GNN}$ simultaneously leverages meta-paths and meta-graphs for rich semantics. 
Moreover, HAE$_{GNN}$ incorporates the content-based interactions between first-order as well as higher-order neighboring nodes. 
The proposed model outperforms state-of-the-art baselines in node classification, node clustering, and visualization tasks.
Experimental results verify the effectiveness of our semantic learning approaches as well as our higher-order attribute-enhancing strategy, and also demonstrate the efficiency as well as the good explainability of the results and proposed model.

There are some potential improvements to the proposed model that could be addressed in the future.
For example, in addition to incorporating the heterogeneity of the nodes, effective approaches to leverage the heterogeneity of the edges could also be explored. 
Also, we may leverage Transformer \cite{vaswani2017attention,hu2020heterogeneous} to explore the nodes' and edges' sequence information. 
Even, we may extend the proposed model to inductive learning in dynamics.
Moreover, as real-world graphs such as social networks contain the information of different modalities, another particularly meaningful research direction would be extending our approach to incorporate information of more modalities such as images.

\section*{Acknowledgement}
This work is supported by the NSFC program (No. U20B2053 and 62002007), 
S\&T Program of Hebei through grant 20310101D, and SKLSDE-2020ZX-12.
Philip S. Yu is supported by NSF under grants III-1763325, III-1909323, and SaTC-1930941.



    
    
    

\bibliography{ref}

\vspace{-15pt}
\begin{IEEEbiography}
[{\includegraphics[width=1in,height=1.25in,clip,keepaspectratio]{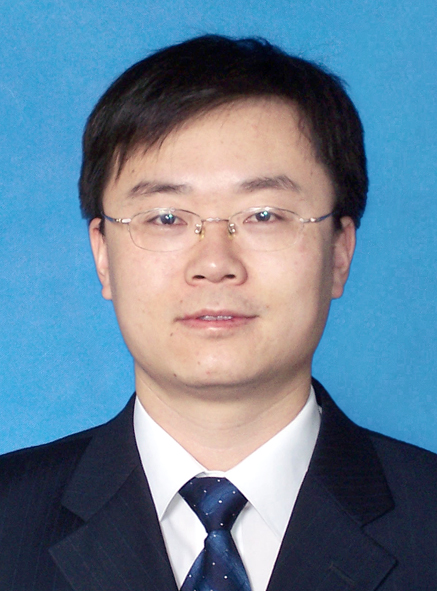}}]
{Jianxin Li} is currently a Professor with the State Key Laboratory of Software Development Environment, and Beijing Advanced Innovation Center for Big Data and Brain Computing in Beihang University. His current research interests include machine learning, distributed system, trust management and network security.
\end{IEEEbiography}

\vspace{-15pt}
\begin{IEEEbiography}[{\includegraphics[width=1.1in,height=1.4in,clip,keepaspectratio]{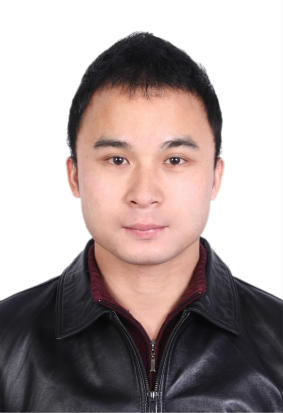}}]{Hao Peng} is currently an assistant professor with Beijing Advanced Innovation Center for Big Data and Brain Computing in Beihang University, and School of Cyber Science and Technology in Beihang University. His research interests include representation learning, text mining and social network mining.
\end{IEEEbiography}

\vspace{-15pt}
\begin{IEEEbiography}[{\includegraphics[width=1in,height=1.25in,clip,keepaspectratio]{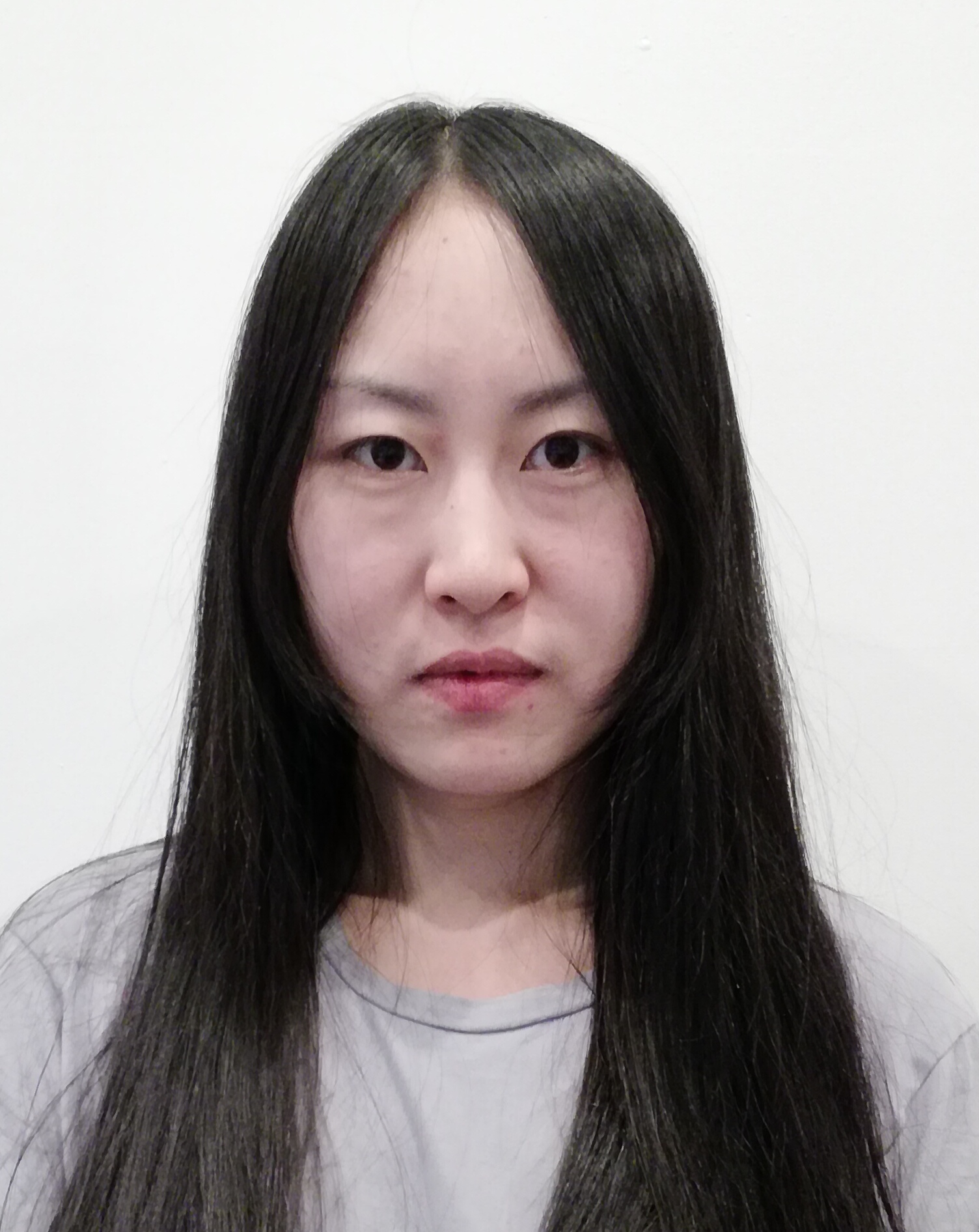}}]{Yuwei Cao} is currently a Ph.D. candidate in the Department of Computer Science at University of Illinois at Chicago. Her research interests include representation learning, graph embedding and social network mining.
\end{IEEEbiography}

\vspace{-15pt}
\begin{IEEEbiography}[{\includegraphics[width=1in,height=1.25in,clip,keepaspectratio]{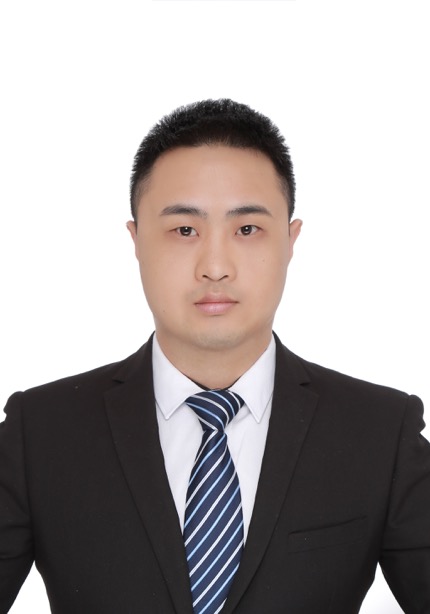}}]{Yingtong Dou} is currently a Ph.D. candidate in the Department of Computer Science at University of Illinois at Chicago. His research interests include graph mining, fraud detection and secure machine learning.
\end{IEEEbiography}

\vspace{-15pt}
\begin{IEEEbiography}[{\includegraphics[width=1in,height=1.25in,clip,keepaspectratio]{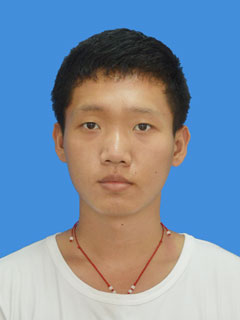}}]{Hekai Zhang} is a graduate student in the School of Information Science and Engineering, Yanshan University. 	His main research interests include machine learning and Graph neural network.
\end{IEEEbiography}

\vspace{-15pt}
\begin{IEEEbiography}[{\includegraphics[width=1in,height=1.25in,clip,keepaspectratio]{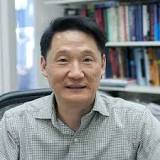}}]{Philip S. Yu} is a Distinguished Professor and the Wexler Chair in Information Technology at the Department of Computer Science, University of Illinois at Chicago. 
He is a Fellow of the ACM and IEEE. 
Dr. Yu was the Editor-in-Chiefs of ACM Transactions on Knowledge Discovery from Data (2011-2017) and IEEE Transactions on Knowledge and Data Engineering (2001-2004).
\end{IEEEbiography}

\vspace{-15pt}
\begin{IEEEbiography}
[{\includegraphics[width=1in,height=1.2in,clip,keepaspectratio]{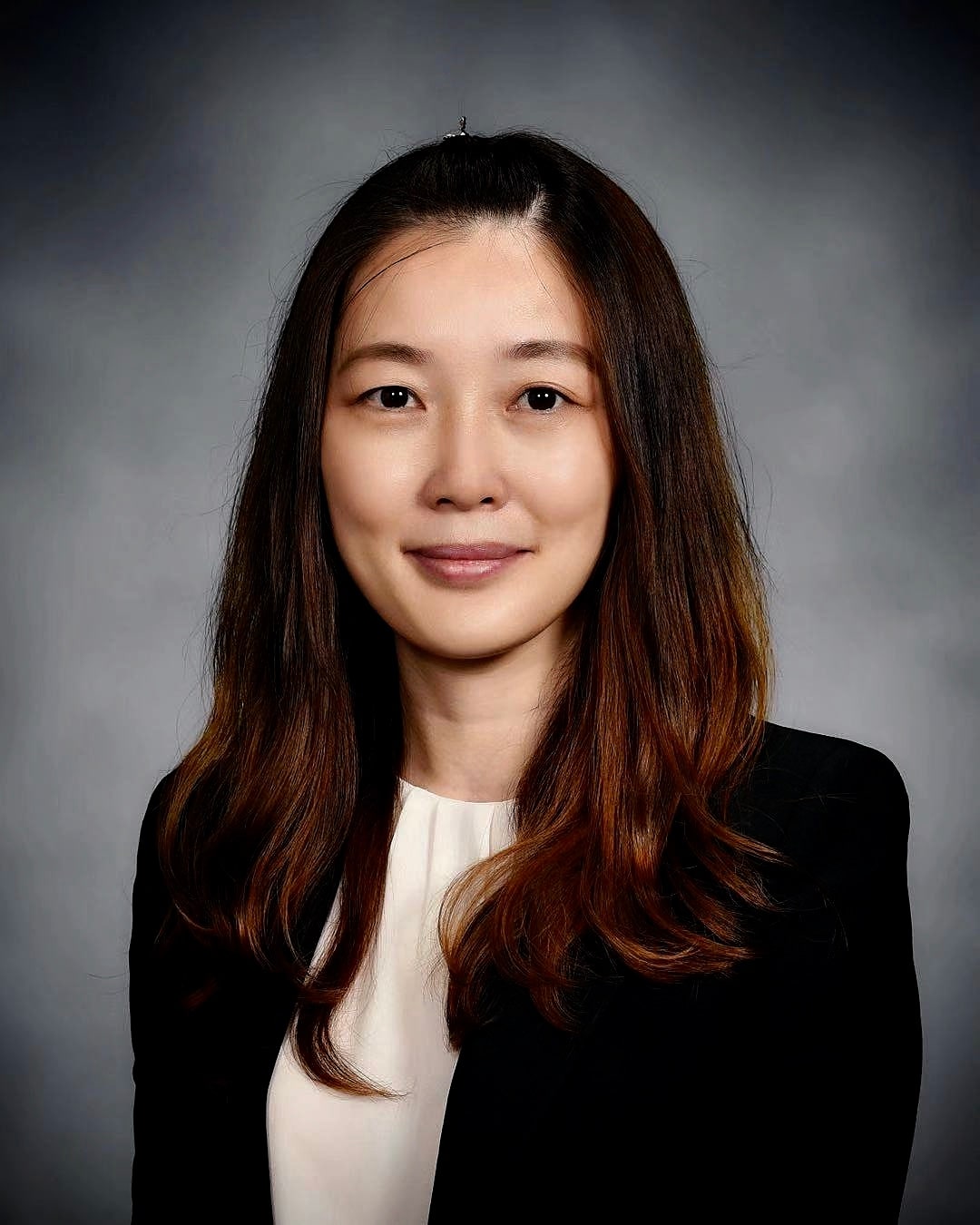}}]
{Lifang He} is currently an Assistant Professor in the Department of Computer Science and Engineering at Lehigh University. Before her current position, Dr. He worked as a postdoctoral researcher in the Department of Biostatistics and Epidemiology at the University of Pennsylvania. Her current research interests include machine learning, data mining, tensor analysis, with major applications in biomedical data and neuroscience.
\end{IEEEbiography}


\end{document}